# PlantSegNeRF: A few-shot, cross-species method for plant 3D instance point cloud reconstruction via joint-channel NeRF with multi-view image instance matching

Xin Yang[a,b], Ruiming Du[c], Hanyang Huang[a,b], Jiayang Xie[a,b], Pengyao Xie[a,b], Leisen Fang[a,b], Ziyue Guo[a,b], Nanjun Jiang[d], Yu Jiang[e], Haiyan Cen[a,b,*]

[a] *State Key Laboratory for Vegetation Structure, Function and Construction (VegLab), College of Biosystems Engineering and Food Science, Zhejiang University, Hangzhou 310058, P.R. China*

[b] *Key Laboratory of Spectroscopy Sensing, Ministry of Agriculture and Rural Affairs, Hangzhou 310058, P.R. China*

[c] *Department of Biological and Environmental Engineering, Cornell University, Ithaca, NY 14850*

[d] *Amway (China) Botanical R&D Center, Wuxi 214115, P.R. China*

[e] *Horticulture Section, School of Integrative Plant Science, Cornell AgriTech, Geneva, NY 14456*

## Abstract

Organ segmentation of plant point clouds is a prerequisite for the high-resolution and accurate extraction of organ-level phenotypic traits. Although the fast development of deep learning has boosted much research on segmentation of plant point clouds, the existing techniques for organ segmentation still face limitations in resolution, segmentation accuracy, and generalizability across various plant species. In this study, we proposed a novel approach called plant segmentation neural radiance fields (PlantSegNeRF), aiming to directly generate high-precision instance point clouds from multi-view RGB image sequences for a wide range of plant species. PlantSegNeRF performed two-dimensional (2D) instance segmentation on the multi-view images to generate instance masks for each organ with a corresponding instance identification (ID). The multi-view instance IDs corresponding to the same plant organ were then matched and refined using a specially designed instance matching (IM) module. The instance NeRF was developed to render an implicit scene containing color, density, semantic and instance information, which was ultimately converted into high-precision plant instance point clouds based on volume density. The results proved that in semantic segmentation of point clouds, PlantSegNeRF outperformed the commonly used methods, demonstrating an average improvement of 16.1%, 18.3%, 17.8%, and 24.2% in precision, recall, F1-score, and intersection over union (IoU) compared to the second-best results on structurally complex datasets. More importantly, PlantSegNeRF exhibited significant advantages in instance segmentation. Across all plant datasets, it achieved average improvements of 11.7%, 38.2%, 32.2% and 25.3% in mean precision (mPrec), mean recall (mRec), mean coverage (mCov), and mean weighted coverage (mWCov), respectively. Furthermore, PlantSegNeRF demonstrates superior few-shot, cross-species performance, requiring only multi-view images of few plants to train models applicable to specific or similar varieties. This study extends organ-level plant phenotyping and provides a high-throughput way to supply high-quality 3D data for developing large-scale artificial intelligence (AI) models in plant science.

# 1 Introduction

High-throughput plant phenotyping, which can provide detailed and efficient plant traits from cellular tissues to canopy populations, is vital for accelerating crop breeding, optimizing agricultural practices, and deepening ecological studies with a better understanding of the interactions among genotype (G), environment (E), and management (M) (Sun et al., 2022). The extraction of organ-level phenotypic traits is particularly desirable for developing crop growth models, designing ideotypes, and conducting upscaling studies from organs to canopies (Jin et al., 2021; Miao et al., 2021). Plant organ segmentation is a prerequisite for high-resolution and accurate plant organ phenotyping, especially in the three-dimensional spatial domain, and has become the main challenge due to diverse plant architecture and data scarcity (Ao et al., 2022; Yang et al., 2024b).

With the advancement of computer vision and optical sensing technologies, researchers have developed various methods to perform plant organ analysis. Initially, two-dimensional (2D) imaging techniques have been widely used for organ segmentation, leaf counting and color analysis (Bhagat et al., 2022; Praveen Kumar and Domnic, 2019). However, the single 2D image is susceptible to perturbation from occlusion and lighting conditions, and cannot fully capture the spatial characteristics of plants (Luo et al., 2023; Yan et al., 2024). In contrast, emerging 3D sensing technologies, such as X-ray computed tomography (CT), structured light sensors, laser scanners, LiDARs, and time-of-flight (ToF) cameras, enable the construction of comprehensive 3D plant point cloud model that better represent spatial structures (He et al., 2025; Yuan et al., 2019). Although intensive studies have been reported, these 3D imaging approaches often face challenges such as high costs, limited portability, or sensitivity to ambient light. Furthermore, their generalizability to complex application scenarios and diverse plant organs is still limited.

Photogrammetry-based 3D reconstruction has gained increasing popularity in recent years because of the use of affordable sensors such as RGB cameras and the potential to diverse applications (Farshian et al., 2023). Conventional photogrammetry-based 3D reconstruction begins by using image features to estimate camera poses and establish image correspondences through techniques such as structure from motion (SfM). This is followed by multi-view stereo (MVS) to reconstruct dense 3D representations and visual renderings. Combining multi-view imagery and conventional photogrammetry-based 3D reconstruction can obtain high-precision plant point clouds for organ segmentation. While various MVS algorithms have steadily improved the quality of 3D reconstruction and rendering for regular geometric surfaces such as buildings (Schönberger et al., 2016; Schönberger and Frahm, 2016), they often struggle with complex and intricate object architecture such as plants, resulting in fragmented reconstructions and significant noise artifacts (Hu et al., 2024). Recent studies have focused on using other representations (e.g., neural implicit representations and Gaussian splatting) to improve the quality of reconstructing fine-grained details and complex scenes. Notable efforts include neural representation of signed distance function (SDF) (Park et al., 2019), neural radiance fields (NeRF) (Mildenhall et al., 2022), and 3D Gaussian splatting (3DGS) (Kerbl et al., 2023). Both neural SDFs and NeRFs are implicit representation of a 3D scene, but they differ in the design purpose. Neural SDFs aim to capture fine geometric details by directly modeling scene surfaces, whereas NeRFs focus on synthesizing novel views with visually pleasing appearances and do not strictly require accurate and detailed geometry. While neural SDFs seem to be well suited to plant phenotyping because of its capability of resolving fine geometric details, NeRFs have been used more frequently as NeRFs are relatively easier to converge in practice (Qiu et al., 2023). NeRF variants in diverse plant phenotyping scenarios were employed for plant phenotyping, such as real-time 3D reconstruction in field environments (Arshad et al., 2024a), large-scale scene generation in strawberry orchards (Zhang et al., 2024), and detailed organ modeling for organ-level phenotyping (Yang et al., 2024a). Unlike neural SDFs and NeRFs, 3DGS uses explicit representation of 3D Gaussians to simultaneously model scene surfaces and colors, and is another cutting-edge approach that has been successfully applied to plant 3D reconstruction including cotton (Jiang et al., 2024), rapeseed (Shen et al., 2025), and wheat (Stuart et al., 2025). Although the 3DGS is capable of producing high-quality explicit point cloud outputs, such as SuGaR (Guédon and Lepetit, 2023) and 2DGS (Huang et al., 2024), the processing time would increase significantly, which limits their applications in high-throughput 3D reconstruction.

Based on the reconstructed 3D plant point cloud, the common next step is the point cloud segmentation, which mainly falls into two categories: feature-based and deep learning-based approaches. Feature-based methods are tailored to the structural characteristics of specific crops. Ma et al. (2023) used skeleton structures to distinguish rapeseed stems from siliques. Jin et al. (2019) applied normalized-vector growth to separate maize stems from leaves. Both methods achieved recall values above 0.92. Nevertheless, as they are designed based on the characteristics of specific crops, their segmentation performance on other crops have not been satisfied yet. For deep learning-based methods, representative segmentation algorithms include PointNet++, dynamic graph convolutional neural network (DGCNN), and position adaptive convolution (PAConv). PointNet and PointNet++ pioneered the use of annotated 3D point cloud data for training and prediction, enhancing the generalizability across different datasets (Qi et al., 2017a, 2017b). While PointNet does not explicitly capture local neighborhoods, PointNet++ addresses this limitation by incorporating hierarchical local features through grouping methods to learn local geometric structures. However, both methods are limited by their reliance on a fixed set of sampling and grouping strategies, which can struggle with highly non-uniform point densities and often miss fine-grained geometric details due to the loss of individual point information during aggregation. Subsequently, DGCNN leveraged the local geometric structures by constructing local neighborhood graphs and applying convolution-like operations on the graph edges (Wang et al., 2019). It demonstrates strong robustness to rotations and non-rigid transformations, but remains sensitive to noise and outliers. Xu et al. (2021a) proposed PAConv, which achieves position-adaptive convolution kernels via learnable weight matrices. However, its multi-branch design leads to significantly increased memory usage. Additionally, several specially designed frameworks by using above basic deep learning architectures were proposed for specific segmentation tasks. Du et al. (2023) proposed plant segmentation transformer (PST) with self-attention mechanisms for rapeseed silique segmentation and achieved the IoU of 93.96%. Li et al. (2022a, 2022b) developed PlantNet and PSegNet by enhancing plant downsampling strategies and network architectures for segmentation, demonstrating cross-species capabilities across tobacco, tomato, and sorghum.

Although existing deep learning-based methods for plant point cloud segmentation have achieved certain successes, they fundamentally face two main challenges. The first challenge is the heavy reliance on extensive finely annotated point cloud data for training. The second lies in insufficient segmentation performance for complex plants, primarily due to the necessity of downsampling to sparse point counts, which hinders networks from capturing intricate structural features. From another perspective, Shi et al. (2019) proposed an alternative strategy, which mapped 2D image segmentation results to 3D space. However, it assigned distinct 2D classification labels (e.g., “leaf 1” and “leaf 2”) to individual leaves, requiring the network to learn discriminative features for separating leaf instances in 2D space. This approach is currently restricted to simple crops with minimal leaf overlap, such as those in the two-leaf stage. Therefore, we proposed a novel approach called plant segmentation neural radiance fields (PlantSegNeRF), aiming to directly generate high-precision instance point clouds from multi-view RGB image sequences for a wide range of plant species. To the best of our knowledge, PlantSegNeRF is the first to perform 2D instance segmentation on images, match instance identifications (IDs) across multiple views, and finally developed an instance NeRF for 3D instance point clouds reconstruction. The main contributions of our work are summarized as follows:

(i) A comprehensive dataset of well-labeled 2D images and point clouds dataset of plants was established, including various varieties and growth stages. 50 plant samples were collected for each type.

(ii) A novel multi-view image instance matching (IM) module was proposed to align plant organ instance IDs across different viewpoints, serving as the foundation for organ-level instance segmentation.

(iii) A multi-channel instance NeRF module with encoding color, semantic, and instance information was developed to achieve high-precision mapping of 2D image colors, semantics, and aligned instances into 3D space, enabling point cloud background removal and fine-grained segmentation of plant organs.

# 2 Methods

## 2.1. Framework

Fig. 1 describes the workflow of the proposed PlantSegNeRF method that enables the 3D reconstruction of instance point clouds of plants from multi-view images. 2D instance segmentation were first performed on the multi-view images to generate instance masks for each organ with a corresponding instance identification (ID). The multi-view instance IDs corresponding to the same plant organ were then matched and refined using a specially designed instance matching module. The instance NeRF was developed based on the original 2D images, matched instance information, semantic information, and camera intrinsic and extrinsic parameters to render an implicit scene, containing color, density, semantic and instance information. The implicit scene was ultimately converted into instance plant point clouds based on the volume density. Additionally, the pipeline for semantic point cloud reconstruction without instance information is presented in Fig. 1d, which serves dual purposes: providing a streamlined approach for applications requiring only semantic segmentation, and combining with spatial clustering as a baseline to evaluate the effectiveness of our instance matching module.

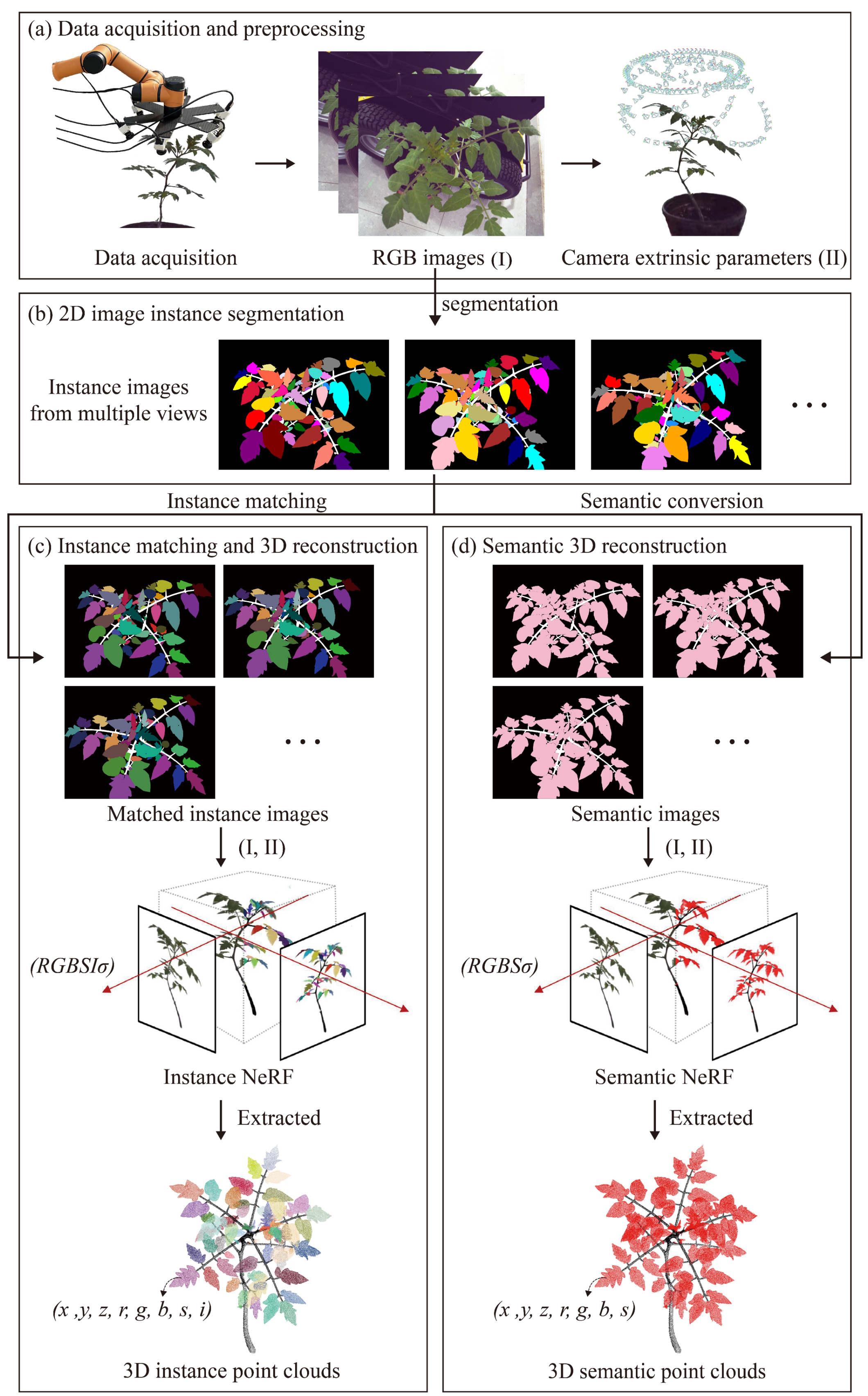

**Fig. 1.** Workflow of the plant segmentation neural radiance fields (PlantSegNeRF) method: (a) data acquisition and preprocessing, (b) 2D image instance segmentation, (c) instance matching and 3D reconstruction, (d) semantic 3D reconstruction.

### 2.2 2D image segmentation and instance matching module

The You Only Look Once version 11 (YOLOv11) was implemented for 2D image instance segmentation to obtain pixel-wise semantic and instance information (Jocher and Qiu, 2023). A bidirectional-voting-scheme-based instance matching (IM) module was developed to unify instance IDs across multi-view 2D images as shown in Fig. 2. The masked images with initial instance labels were analyzed to identify the frequency distribution of instance counts, from which a main image was randomly selected among those with the highest frequency. An adaptive resolution minimum rectangle sampling strategy (detailed methodology described in Equation S1) was then applied to the instances in the main image, ensuring sufficient sampling density for small instances while maintaining computational efficiency for larger ones. Sampling points were projected into the 3D point cloud space and then to auxiliary images using camera extrinsic parameters to establish a bidirectional voting statistical model. The bidirectional voting mechanism consisted of two complementary processes: forward voting and inverse voting. Forward voting evaluated the correspondence from main image instances to auxiliary image instances by analyzing the distribution of projected sampling points within auxiliary images. For each main image instance, voting tables were established based on the number of projected points falling into different auxiliary instances, background (BG), or outside (OUT) the image boundaries, as illustrated in the left panel of Fig. 2c. In parallel, inverse voting assessed the correspondence from auxiliary image instances back to main image instances. For each auxiliary image instance, the number of projected points within its mask that originated from different main image instances was counted, as shown in Fig. 2c (right). The bidirectional voting results were then integrated to establish robust instance correspondences, where mutual correspondence between forward and reverse voting indicated reliable matches.

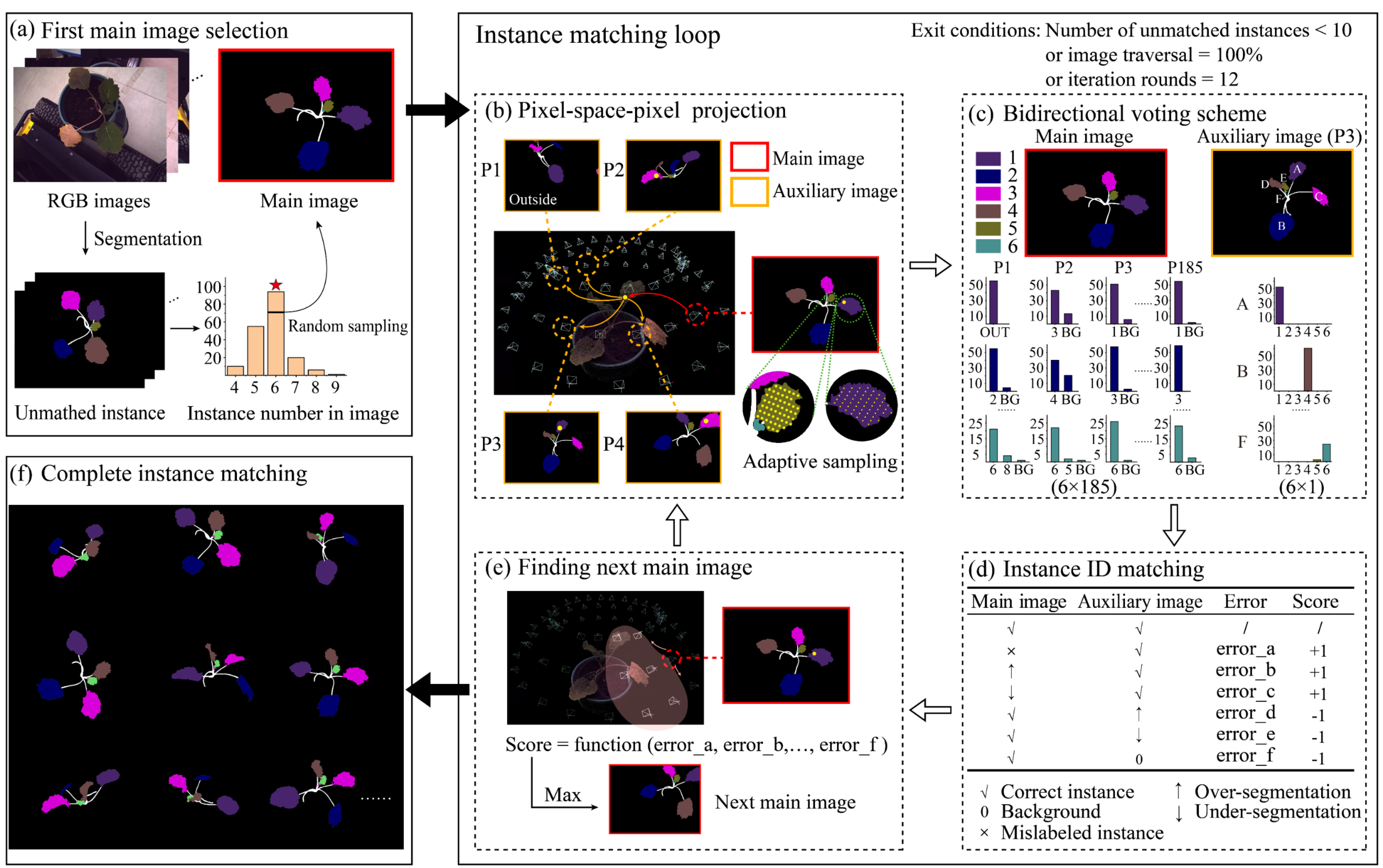

**Fig. 2.** Pipeline for implementing instance matching module for multi-view 2D images: (a) first main image selection, (b) pixel-space-pixel projection, (c) bidirectional voting scheme, where outside (OUT) indicates projection points falling outside the image frame, background (BG) indicates projection points landing in the background. In the 6×185 matrix, 6 represents the number of instances in the main image, and 185 represents the number of auxiliary images. In the 6×1 matrix, 6 represents the number of instances in the Auxiliary image. (d) instance identifications (IDs) matching and problem detection, (e) finding next main image, and (f) complete instance matching.

Based on the bidirectional voting results, instance IDs were matched across different views while abnormal 2D segmentation cases were detected, primarily including six patterns: (a) background mislabeled as instances in main image, (b) over-segmentation in the main image where a single instance is incorrectly divided into multiple regions, (c) under-segmentation in the main image where multiple independent instances are incorrectly merged into a single region, (d) over-segmentation in auxiliary images, (e) under-segmentation in auxiliary images, and (f) instance loss issues. Different processing strategies were developed for various types of errors, including background instance elimination based on multi-view consistency and instance refinement through merging or removal guided by voting distribution features, as detailed in Table 1. The IM module employed a dynamic iterative processing mechanism, utilizing a composite scoring function incorporating the instance quantity and the error label rate to select the optimal adjacent view as the next main image. Iteration terminated when meeting any of the following conditions: completing a full traversal of the image, reaching 12 iterations based on the preliminary test with acceptable processing time, or when the number of unprocessed instances falls below a threshold of 10 (indicating that most instance matching is complete), which ensures that the resulting instance IDs are predominantly consistent across different views. Examples of error handling by the IM module and the number of various types of errors discovered during iterations can be found in Fig. S1.

**Table 1.** Error detection criteria and processing strategies for abnormal 2D segmentation patterns in the bidirectional voting instance matching module

| Error code | Error pattern | Detection criteria | Processing strategy |
|---|---|---|---|
| (a) | Background mislabeled in main image | Forward voting shows main image instance receives highest votes for background regions across >50% of auxiliary images | Instance elimination |
| (b) | Over-segmentation in main image | Forward voting shows multiple main image instances receive highest votes for the same auxiliary instance across >50% of auxiliary images | instance identifications (ID)unification and global label updates |
| (c) | Under-segmentation in main image | Inverse voting shows multiple auxiliary instances receive highest votes for a single main image instance across >50% of auxiliary images | Skip instance assignment to prevent error propagation |
| (d) | Over-segmentation in auxiliary images | Multiple instances in the same auxiliary image receive highest votes for the same main image instance which has no under-segmentation issue | Instance ID unification and global label updates |
| (e) | Under-segmentation in auxiliary images | Single auxiliary instance receives highest votes for multiple main image instances which have no over-segmentation issue | Score penalty in next main image selection |
| (f) | Instance loss | Normal main image instance receives highest votes for background regions in auxiliary images | Score penalty in next main image selection |

### 2.3 Instance NeRF for 3D reconstruction

PlantSegNeRF included instance NeRF and semantic NeRF networks as shown in Fig. 1. While sharing similar characteristics, our work focused on the complex instance NeRF network, which mapped multi-view 2D matched instance images to 3D instance point clouds. The network architecture consisted of modules for image encoding, position and direction encoding, multi-stream MLPs, volume rendering and loss functions, and point clouds extraction and decoding (Fig. 3).

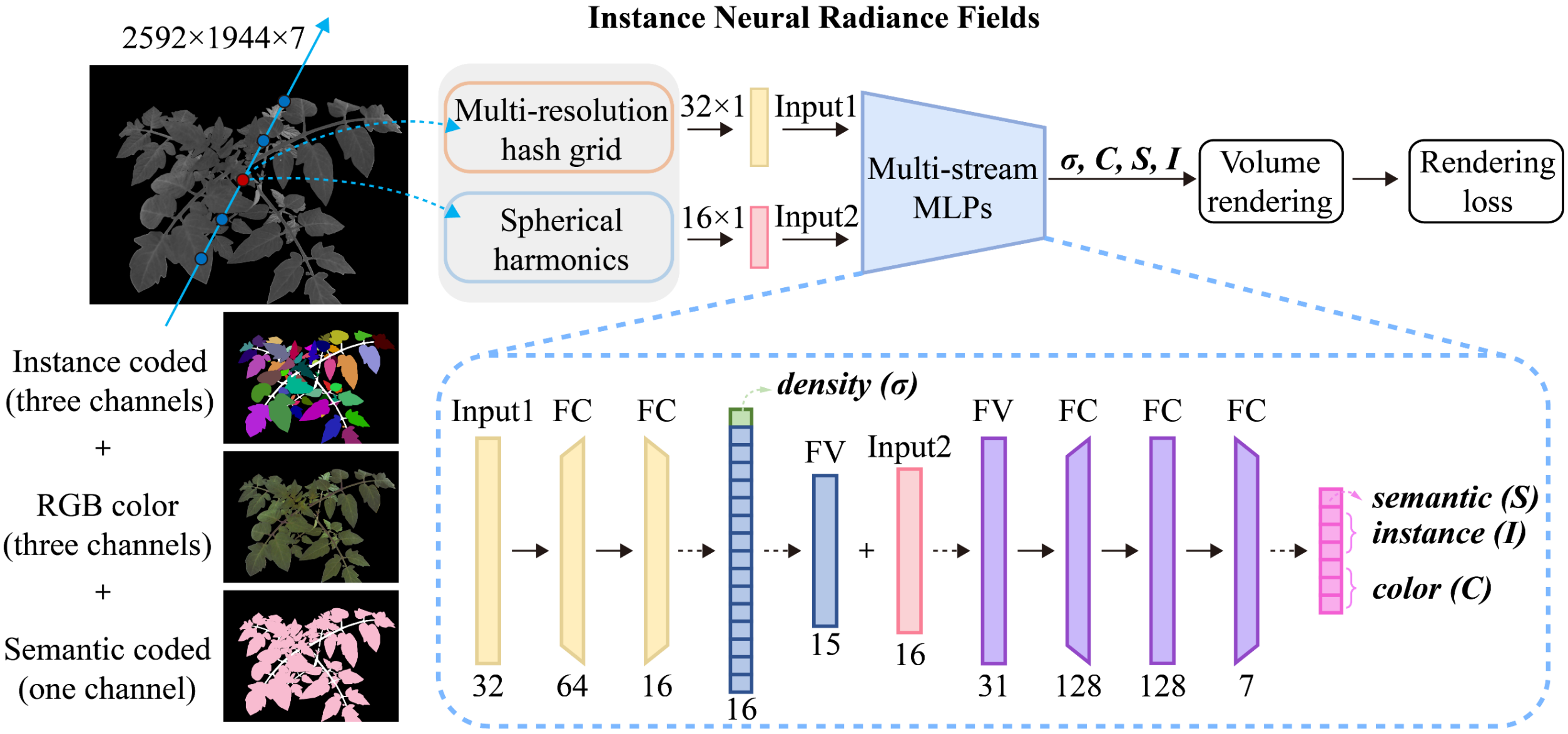

**Fig. 3.** Network architecture for instance neural radiance fields (NeRF), where 2592×1944 represents the pixel resolution of 2D images, 7 indicates the combination of three instance encoding layers, three color encoding layers, and one semantic encoding layer, and density (σ), semantic (S), instance (I), and color (C) are the outputs of multi-stream multilayer perceptrons (MLPs) networks.

#### 2.3.1 Image encoding

For instance and semantic information on 2D images, encoding was implemented to facilitate subsequent loss calculations. Semantic information was encoded using a single 8-bit channel ($S_1$), while instance numbering, being more complex than semantic categorization, was encoded using three 8-bit channels ($I_1$, $I_2$, $I_3$). Color information was preserved in the standard 24-bit RGB format without additional processing. Consequently, beyond coordinates, each pixel is represented by a seven-channel vector: ($R$, $G$, $B$, $S_1$, $I_1$, $I_2$, $I_3$).

### 2.3.2 Position and direction encoding

To capture high-frequency scene details, such as intricate plant textures and edges, the spatial coordinates (X, Y, Z) were initially encoded. PlantSegNeRF employed multi-resolution hash encoding for more efficient input position encoding. Specifically, 16 multi-resolution layers were utilized, with each layer calculating a two-dimensional positional encoding. The concatenation of these 16 layers resulted in a 32-dimensional positional representation. Moreover, to enhance the network capability in capturing variations in surface reflection and illumination due to changes in the viewing direction, spherical harmonic encoding was employed for the directional parameters $\theta$ and $\phi$, as described in Equation (1):

$$Y_j^m(\theta,\phi) = \begin{cases} \sqrt{2}\, K_j^m \cos(m\phi) P_j^m(\cos\theta) \ \ (m > 0) \\ \sqrt{2}\, K_j^m \sin(-m\phi) P_j^{-m}(\cos\theta) \ \ (m < 0) \\ K_j^0 P_j^0(\cos\theta) \ \ (m = 0) \end{cases} \tag{1}$$

where $\theta$ represents the pitch angle, measured from the positive z-axis and ranging from 0 to π, $\phi$ represents the yaw angle, measured from the positive x-axis and ranging from 0 to 2π, and $P_j^m$ denotes the associated Legendre function of degree $j$ and order $m$, defined in Equation (2-4):

$$P_j^m(x) = (-1)^m (1-x^2)^{m/2} \frac{d^m}{dx^m} P_j(x) \tag{2}$$

$$P_n(x) = \frac{1}{2^n n!} \frac{d^n}{dx^n} [(x^2-1)^n] \tag{3}$$

$$K_j^m = \sqrt{\frac{(2j+1)(j-|m|)!}{4\pi(j+|m|)!}} \tag{4}$$

where $j$ and $m$ represent the degree and the order, respectively, both of which are integers, with $m$ ranging from $-j$ to $j$. Increasing values of $j$ facilitate more sophisticated lighting models, resulting in a better approximation to the physical properties of real-world scenarios. However, it would also significantly increase the computational cost during the training process. By considering the trade-off between complexity of the lighting representation and computational efficiency, PlantSegNeRF utilized a third-degree representation, yielding a 16-dimensional feature encoding.

### 2.3.3 Multi-stream MLPs

The multi-stream MLPs had two inputs, including a 32-dimensional spatial position encoding and a 16-dimensional directional encoding. The spatial encoding was initially processed through two fully connected layers, expanding from 32 to 64 dimensions before being reduced to 16 dimensions to extract key spatial features. One of these dimensions was used to represent the volume density, while the remaining 15 dimensions were combined with the directional encoding to form a 31-dimensional feature vector. This combined vector was further processed through multiple fully connected layers. The output of the network was a 7-dimensional feature vector, where the first three dimensions encode the RGB color information, the next three dimensions represent the instance embeddings and the last dimension is related to the semantic information.

### 2.3.4 Volume rendering and loss functions

The primary objective of volume rendering was to project the color, semantic and instance attributes of spatial points onto the image plane, thereby enabling the computation of squared residuals between the rendered output and the ground truth, thus forming an effective loss function. The mathematical formulation for volume rendering is presented in Equation 5-6:

$$C(r) = \int_{t_n}^{t_f} T(t)\,\sigma\big(\boldsymbol{r}(t)\big)\boldsymbol{c}(\boldsymbol{r}(t), \boldsymbol{d})dt, where\ T(t) = exp\left(-\int_{t_n}^{t} \sigma(\boldsymbol{r}(s))ds\right) \tag{5}$$

$$S(r) = \int_{t_n}^{t_f} T(t)\,\sigma\big(\boldsymbol{r}(t)\big)\boldsymbol{s}(\boldsymbol{r}(t), \boldsymbol{d})dt, where\ T(t) = exp\left(-\int_{t_n}^{t} \sigma(\boldsymbol{r}(s))ds\right) \tag{6}$$

$$I(r) = \int_{t_n}^{t_f} T(t)\,\sigma\big(\boldsymbol{r}(t)\big)\boldsymbol{i}(\boldsymbol{r}(t),\boldsymbol{d})dt, where\ T(t) = exp\left(-\int_{t_n}^{t} \sigma(\boldsymbol{r}(s))ds\right) \tag{7}$$

where $C(r)$, $S(r)$ and $I(r)$ denote the color, semantic and instance information, respectively, $\boldsymbol{r}(t)$ represents points in 3D space, $\boldsymbol{d}$ represents the viewing direction, $t_n$ and $t_f$ represent the near and far boundaries of the 3D scene, $\boldsymbol{c}(\boldsymbol{r}(t),\boldsymbol{d})$, $\boldsymbol{l}(\boldsymbol{r}(t),\boldsymbol{d})$ and $\boldsymbol{i}(\boldsymbol{r}(t),\boldsymbol{d})$ represent the color and semantic value observed from direction at point, $\sigma\big(\boldsymbol{r}(t)\big)$ represents volume density function, describing the ability of physical materials to absorb light, and $T(t)$ represents the cumulative transmittance along the ray from $t_n$ to $t_f$.

The loss function was formulated as the sum of the squared differences between the actual and rendered outputs, encompassing both color, semantic and instance values. The network parameters were iteratively updated using backpropagation and a gradient descent strategy, allowing the model to effectively fit the plant scene. Additionally, the loss function incorporated a coarse-to-fine sampling strategy (detailed parameters shown in Table S1), resulting in both coarse and fine loss components, as expressed in Equation (8):

$$Loss = \sum_{r \in \mathcal{R}} \begin{bmatrix} \left\|\hat{C}_c(r) - C(r)\right\|_2^2 + \left\|\hat{C}_f(r) - C(r)\right\|_2^2 + \left\|\hat{I}_c(r) - I(r)\right\|_2^2 \\ +\left\|\hat{I}_f(r) - I(r)\right\|_2^2 + \left|\hat{S}_c(r) - S(r)\right|^2 + \left|\hat{S}_f(r) - S(r)\right|^2 \end{bmatrix} \tag{8}$$

where $\hat{C}_c(r)$ and $\hat{C}_f(r)$ denote the predicted colors at coarse and fine stages, respectively, while $\hat{I}_c(r), \hat{I}_f(r)$ and $\hat{S}_c(r)$, $\hat{S}_f(r)$ represent the instance and semantic predictions at respective stages, $r$ represents an individual sampled ray, $\mathcal{R}$ represents the set of all sampled rays, $C$ is a three-dimensional vector, $S$ is a one-dimensional scalar, the notation $\|X\|_2^2$ represents the sum of the squared components of the vector.

### 2.3.5 Point clouds extraction and decoding

The marching cubes algorithm was employed to convert implicit NeRF scene into explicit point clouds through threshold-based cube partitioning and linear interpolation. During the extraction process, noise filtering was performed in the neighborhood regions along camera orientations to eliminate sparse noise artifacts commonly observed near camera positions. The extracted point clouds contain spatial coordinates (X, Y, Z), RGB color values, semantic encoding ($S_1$) and instance encoding ($I_1$, $I_2$, $I_3$). Following the 2D image coding protocol, $S_1$ and $I_1$, $I_2$, $I_3$ were decoded to obtain semantic and instance information, yielding a comprehensive point cloud representation with eight-dimensional features ($X$, $Y$, $Z$, $R$, $G$, $B$, $S$, $I$) per point.

# 3 Experiment

### 3.1 Plant dataset preparation

Fig. 4 presents detailed dataset information of six plant species used in this study. These datasets were selected for three main plant segmentation tasks to validate the generalization of PlantSegNeRF across different segmentation scenarios, including stem-leaf segmentation, yield-organ extraction, and multi-organ segmentation (Li et al., 2025; Wu et al., 2022). For stem-leaf segmentation, four plant species were selected based on their distinct plant architectures to evaluate robustness of PlantSegNeRF across diverse morphological features. For yield-organ extraction, fruit-stage eggplants were selected to validate the robustness in complex fruit extraction scenarios with low color contrast, severe occlusion, and morphological diversity. To verify the capability in multi-organ segmentation, mutant rapeseed at the podding stage was selected as it exhibits stems, leaves, flowers and siliques simultaneously.

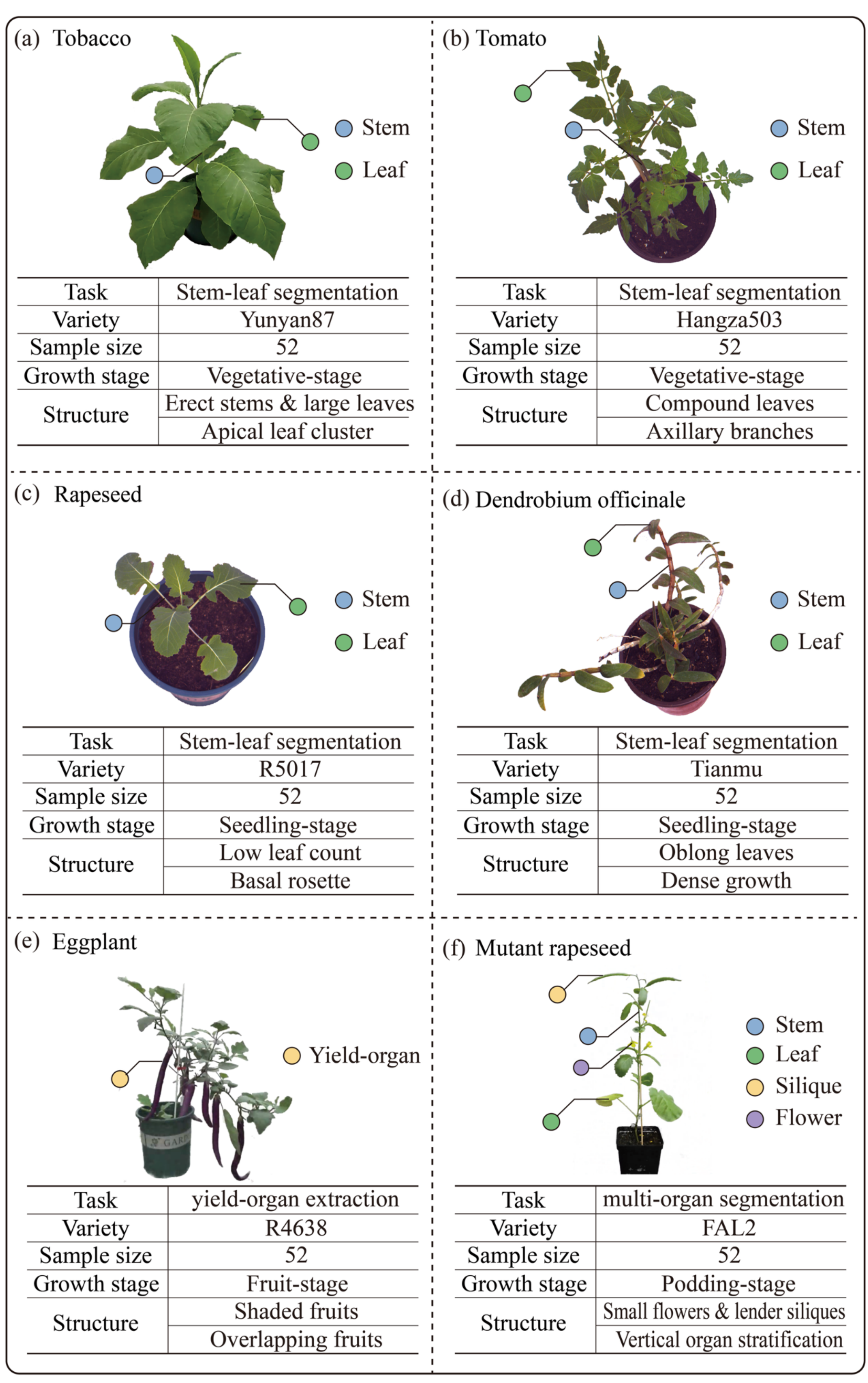

**Fig. 4.** The description of six plant species datasets, including (a) tobacco, (b) tomato, (c) rapeseed, (d) dendrobium officinale, (e) eggplant, and (f) mutant rapeseed.

As shown in Fig. 1, a robotic-arm-driven multi-camera synchronous system was developed for rapid acquisition of multi-view images from different plants. Six cameras were connected to a small host computer for synchronized data collection. A total of 186 images per plant sample from multiple viewpoints were acquired within 30-60 seconds based on the complexity of plant architectures, and the image size was 2595×1944 pixels with auto exposure time. For dataset construction, 52 plant samples for each species were used for multi-view image acquisition, with two samples for 2D instance segmentation training and the remaining 50 samples for performance evaluation of PlantSegNeRF. Only 120 manual annotated images from two training plants with 60 of each were needed for 2D instance segmentation training, constituting a few-shot approach with minimal plant sample requirement. A one-time estimation of camera extrinsic matrices was achieved through structure from motion (SfM) algorithm (Schönberger and Frahm, 2016), which was used across all samples within the same group, benefiting from the high positioning accuracy and repeatability of the robotic arm.

### 3.2 Model training

To validate the versatility of our methodology, independent annotation and training were performed for each plant dataset. The 120 annotated images were proportionally resized so that their longer side was 2160 pixels and were subsequently split into training and validation sets at an 8:2 ratio. The YOLOv11 model was trained and evaluated using the same image size throughout the process.

For the training of the plant instance NeRF model, a 9:1 split of images were employed for training and testing, respectively, with a total of 10,000 training iterations. The model operated directly on the original image size without any downsampling. During each iteration, a batch of 2048 rays were processed, and point cloud extraction was set to sample 1,000,000 points. Detailed configurations for both the YOLOv11 and instance NeRF models and the loss function plots of various NeRF loss components are provided in Table S1 and Fig. S2, respectively.

### 3.3 Evaluation metrics

Precision, recall, F1-score, and IoU were used to assess semantic segmentation accuracy. For all four measures, the higher value means better segmentation. For each semantic class, IoU is a standard intersection over union representation. Precision reflects the proportion of points correctly classified by the network to the total predicted points of a semantic class. Recall is the proportion of correctly predicted points in a semantic class by the network to the total number of ground truth points in the semantic class. F1-score is a harmonic average of precision and recall. These four semantic measures are defined as follows:

$$Precision = \frac{TP}{TP + FP} \tag{9}$$

$$Recall = \frac{TP}{TP + FN} \tag{10}$$

$$F1 = 2 \cdot \frac{Precision \cdot Recall}{Precision + Recall} \tag{11}$$

$$IoU = \frac{TP}{TP + FP - FN} \tag{12}$$

where $TP$, $FP$ and $FN$ are the number of true positive, false positive, and false negative points of a semantic class, respectively.

Mean precision (mPrec), mean recall (mRec), mean coverage (mCov), and mean weighted coverage (mWCov) were used to evaluate the performance of instance segmentation. mCov is defined as the average point-level IoU of instance prediction matched with ground truth and mWCov is a weighted version of mCov. The four instance measures are defined as follows:

$$mPrec = \frac{TP^{ins}}{|O|} \tag{13}$$

$$mRec = \frac{TP^{ins}}{|R|} \tag{14}$$

$$mCov = \frac{1}{R}\sum_{i=1}^{|R|} \max_{j} \text{IoU}\left(P_i^R, P_j^O\right) \tag{15}$$

$$mWCov = \sum_{i=1}^{|R|} w_i \max_{j} \text{IoU}\left(P_i^R, P_j^O\right)$$

$$w_i = \frac{\left|P_i^R\right|}{\sum_k \left|P_k^R\right|} \tag{16}$$

where $TP^{ins}$ is the number of predicted instances having an $IoU$ larger than 0.5 with the ground truth, $|R|$ and $|O|$ are the number of all instances in the ground truth and prediction, respectively. $P_i^R$ is the number of points in the $i$-th ground truth instance and $P_j^O$ is the number of points in the $j$-th predicted instance.

To assess the effectiveness of the IM module, we conducted experiments by removing IM from the pipeline. 3D point clouds with semantic information were generated by using the semantic conversion as shown in Fig. 1. For each semantic class, spatial clustering was performed using the density-based spatial clustering of applications with noise (DBSCAN) to obtain instance information (Ester et al., 1996). The neighborhood radius and minimum sample number of DBSCAN were optimized by the gradient testing for each dataset to achieve the best clustering results. The clustering-based instance segmentation approach was compared with the IM module using four metrics, including mPrec, mRec, mCov, and mWCov.

To assess the point clouds segmentation performance of PlantSegNeRF, we generated 3D point clouds with semantic and instance information on six plant species datasets. The point clouds were then visualized in CloudCompare with both position and color information, followed by manual creation of ground truth semantic and instance labels. Besides, we compared our method with commonly used point cloud segmentation methods, including PointNet++, DGCNN (graph-based), PAConv (adaptive convolution), PlantNet and PSegNet (networks capable of both semantic and instance segmentation). 50 point-clouds with ground truth semantic and instance labels per dataset were split into the training and testing datasets with the ratio of 4:1. Five-fold cross-validation was implemented to ensure statistical reliability.

**3.4 Investigation of multi-view image number impact on PlantSegNeRF performance**

The number of multi-view images has a substantial impact on the performance of both the instance matching and NeRF module training. To identify the optimal number of images that balances reconstruction quality and computational efficiency, experiments utilizing varying numbers of images were carried out. Five plant samples were randomly selected from each dataset and tests were performed with 12 different image quantities ranging from 10 to 170. The evaluation metrics included point cloud completeness, IoU for semantic segmentation, mWCov for instance segmentation, and time consumption from data acquisition to final point cloud generation, with the computational devices listed in Table S2. Point cloud completeness evaluation was conducted using reference point clouds as ground truth. Previous research by Arshad et al. demonstrated that nerfacto method outperforms other NeRF-based methods in plant 3D reconstruction accuracy (Arshad et al., 2024b). Based on this finding, reference point clouds were generated using nerfacto with complete image sets for each dataset. The quantitative metric for completeness assessment is defined as follows:

$$Completeness = \frac{1}{|P_{gt}|} \sum_{p \in P_{gt}} \mathbb{I}\left( \min_{q \in P_{test}} \|p - q\| \leq \epsilon \right) \tag{17}$$

where $P_{gt}$ denotes the ground truth point cloud set, $P_{test}$ represents the point cloud set to be evaluated, $\epsilon$ is the distance threshold set to 0.025 in this experiment. $\mathbb{I}(\,)$ is an indicator function that equals 1 when the condition is satisfied and 0 otherwise and $\|\,\|$ represents the Euclidean distance.

# 4 Results

## 4.1 Semantic segmentation of point clouds

PlantSegNeRF exhibited remarkable performance across six plant species datasets, achieving precision, recall, F1-score, and IoU of 95%, 90%, 93% and 88%, respectively (Fig. 5). On the structurally complex datasets of tomato, dendrobium officinale, eggplant, and mutant rapeseed, PlantSegNeRF outperformed all commonly used methods, demonstrating an average improvement of 16.1%, 18.3%, 17.8%, and 24.2% in precision, recall, F1-score, and IoU compared to the second-best results across these datasets. For the relatively simple rapeseed dataset, PlantSegNeRF still maintained a certain lead, while on the tobacco dataset, its performance was comparable to the second-best method. The performance ranking among the other five methods was inconsistent. PAConv excelled in vegetative-stage tomato segmentation but underperformed for the segmentation of dendrobium officinale. While all models performed adequately on vegetative-stage tobacco with its simple rosette structure, other methods struggled with the challenging scenarios of intertwined dendrobium officinale and eggplants with less distinctive features, where their IoU dropped to 48-70%.

Visualization of semantic segmentation for different datasets are shown in Fig. 6. In datasets with complex stem-leaf structures, including tomato and dendrobium officinale, PlantSegNeRF demonstrated superior performance compared to other methods. Notably, when processing complex regions with intertwined stems and leaves, PointNet++ and DGCNN incorrectly identified drooping leaves as stems in seedling-stage rapeseed with twisted stem sections. For yield-organ extraction, PlantSegNeRF presented a great advantage in fruit extraction. In comparison, PointNet++, DGCNN and PAConv failed to effectively differentiate eggplants, while PlantNet and PSegNet only achieved partial fruit detection. In the multi-organ segmentation task of the podding-stage mutant rapeseed dataset, PlantSegNeRF exhibited outstanding performance in handling both subtle features like scattered flowers and complex plant architectures with multi-shape organs.

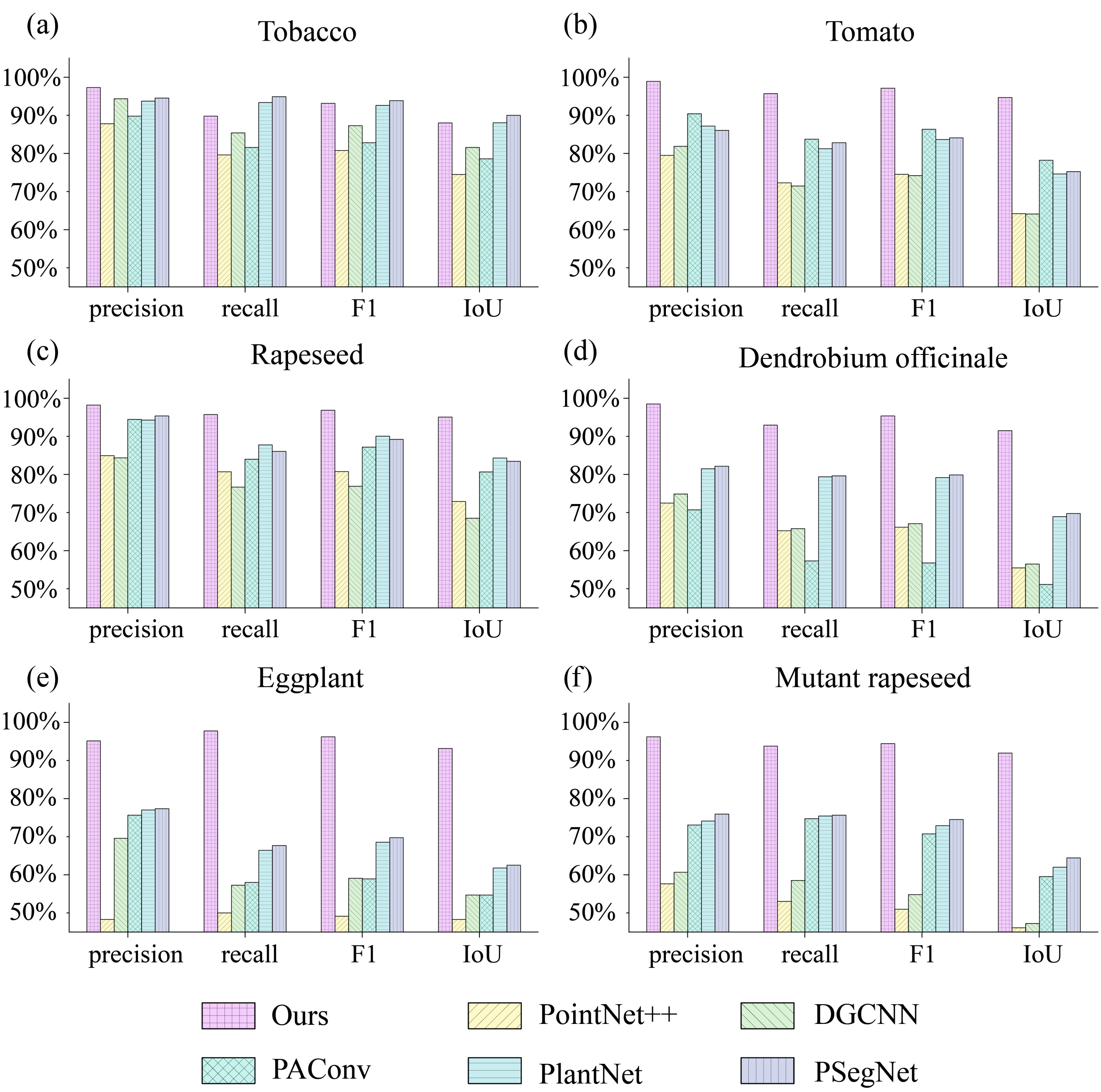

**Fig. 5.** Performance of semantic segmentation with the proposed PlantSegNeRF as well as PointNet++, DGCNN, PAConv, PlantNet, and PSegNet for six plant species datasets: (a) tobacco, (b) tomato, (c) rapeseed, (d) dendrobium officinale, (e) eggplant, and (f) mutant rapeseed.

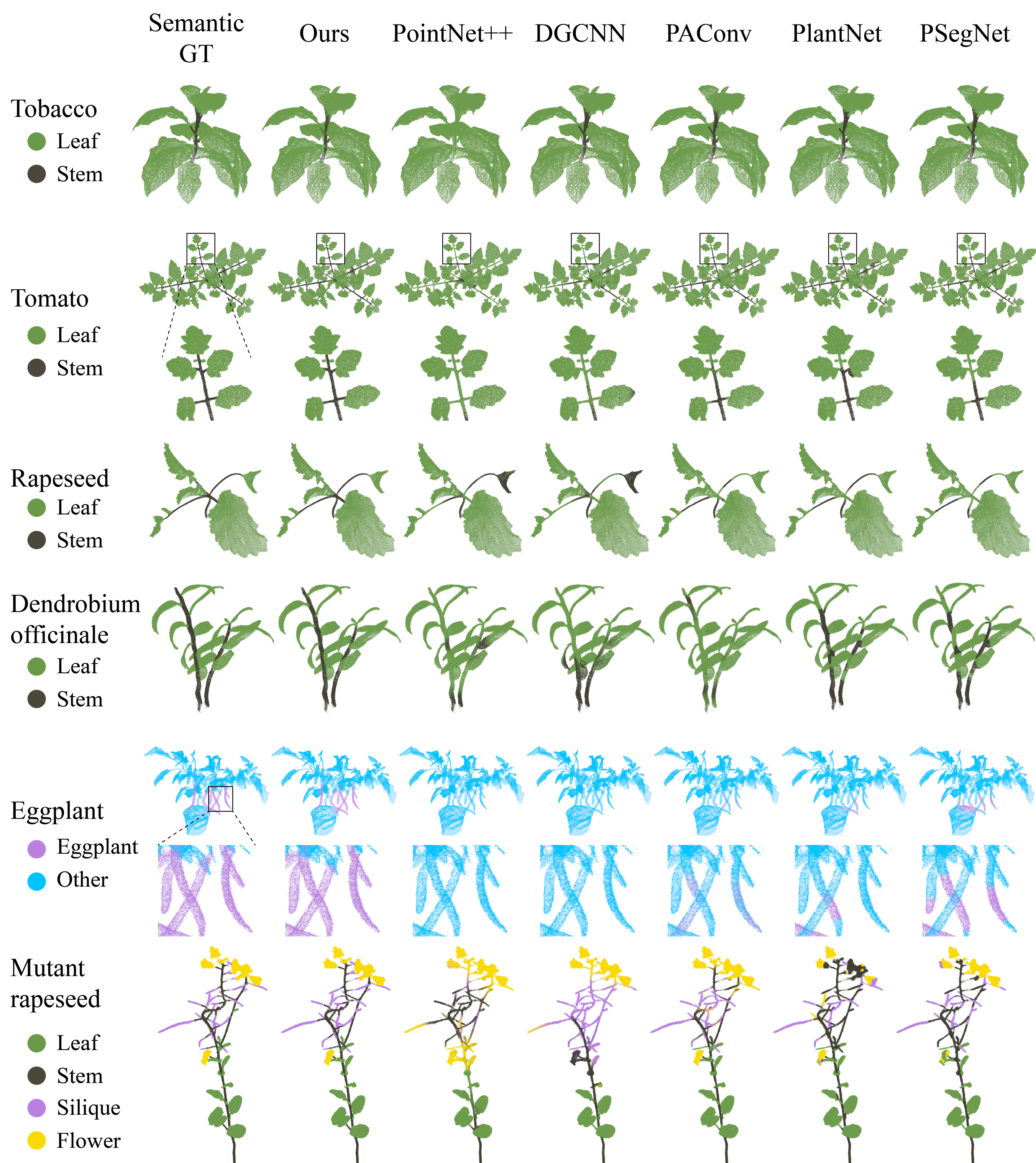

**Fig. 6.** Visualization of the semantic segmentation results for six plant species datasets using the proposed PlantSegNeRF as well as PointNet++, DGCNN, PAConv, PlantNet, and PSegNet.

### 4.2 Instance segmentation of point clouds

Fig. 7 shows that our proposed method PlantSegNeRF exhibited significant advantages in plant point cloud instance segmentation tasks, particularly showing excellent performance when handling complex plant morphologies. Across all plant datasets, PlantSegNeRF achieved average improvements of 11.7%, 38.2%, 32.2% and 25.3% in mPrec, mRec, mCov and mWCov, respectively, compared to the second-best method, PSegNet. As shown in Fig. 8, other methods exhibited limitations that vary across different segmentation tasks. In stem-leaf segmentation, instance boundaries could be misclassified and close leaves cannot be differentiated. In yield-organ segmentation, they frequently misidentified stem point clouds as fruit instances and produced discontinuous segmentation of fruit structures. Furthermore, in multi-object segmentation of rapeseed flowers, they showed limited capability in distinguishing small instances, resulting in simultaneous under-segmentation and over-segmentation issues. In contrast, PlantSegNeRF achieved significant improvements in challenging scenarios including leaf adjacency regions and heterogeneous organ boundaries.

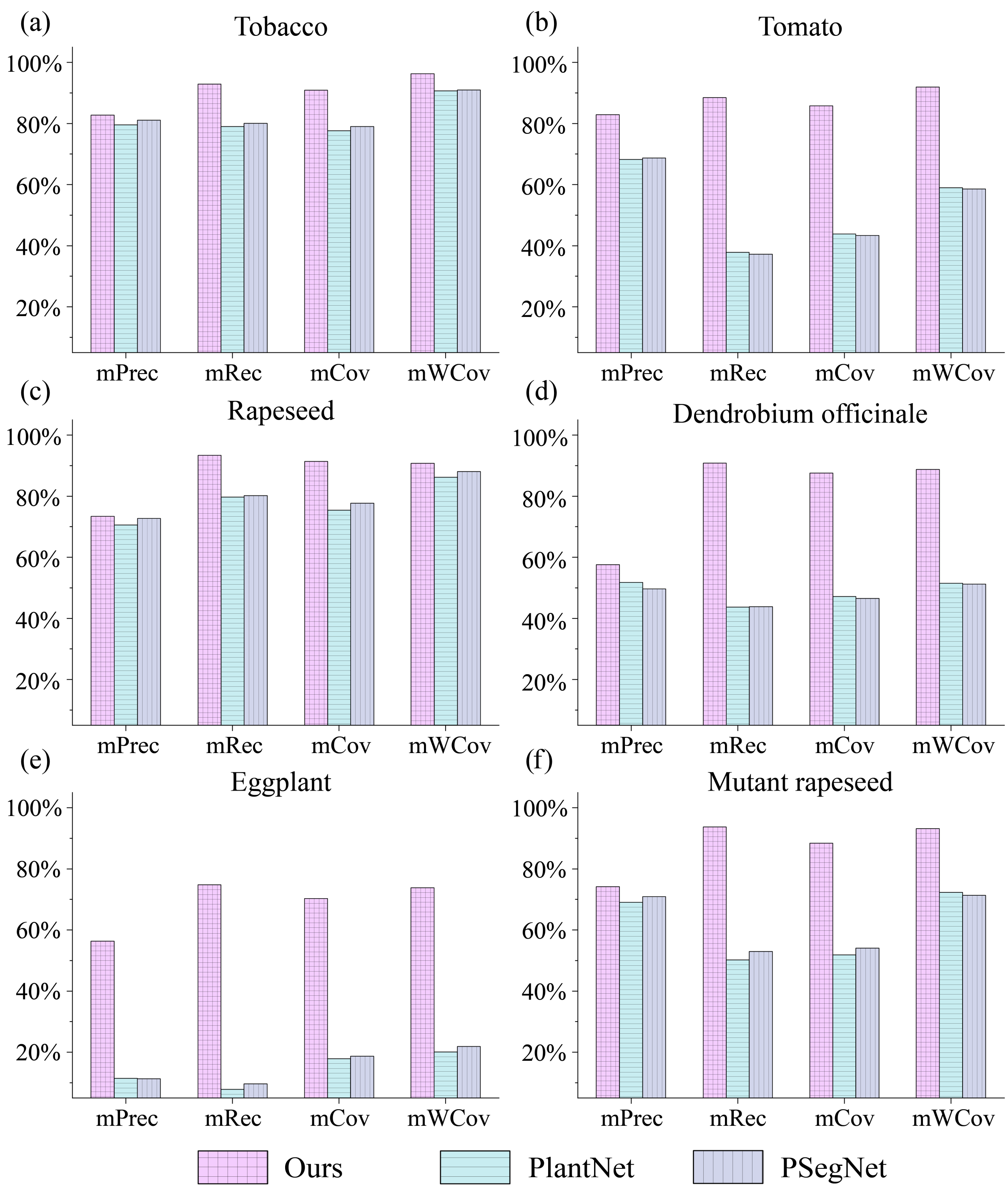

**Fig. 7.** Performance of instance segmentation with the proposed PlantSegNeRF as well as PlantNet and PSegNet for six plant species datasets: (a) tobacco, (b) tomato, (c) rapeseed, (d) dendrobium officinale, (e) eggplant, and (f) mutant rapeseed.

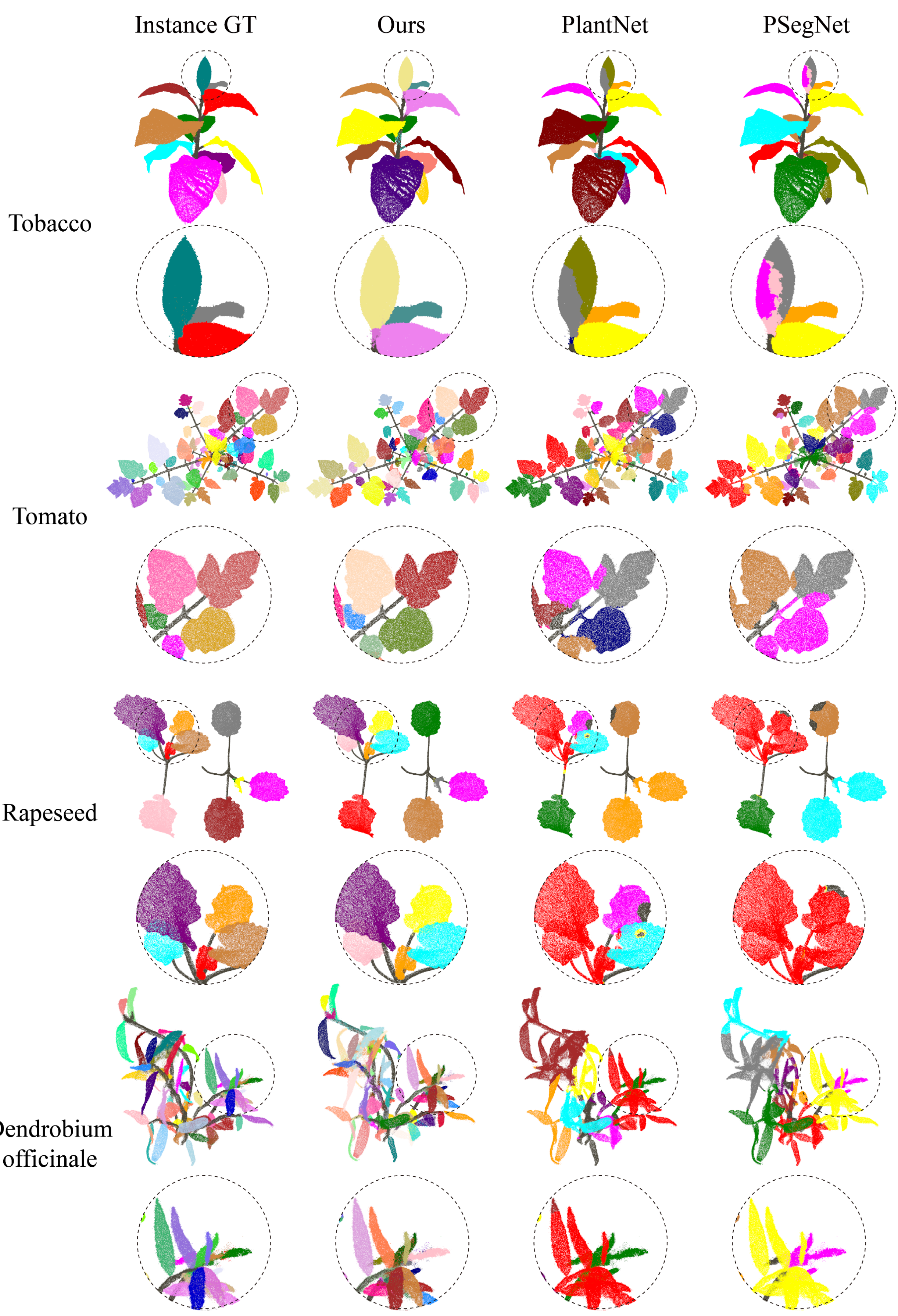

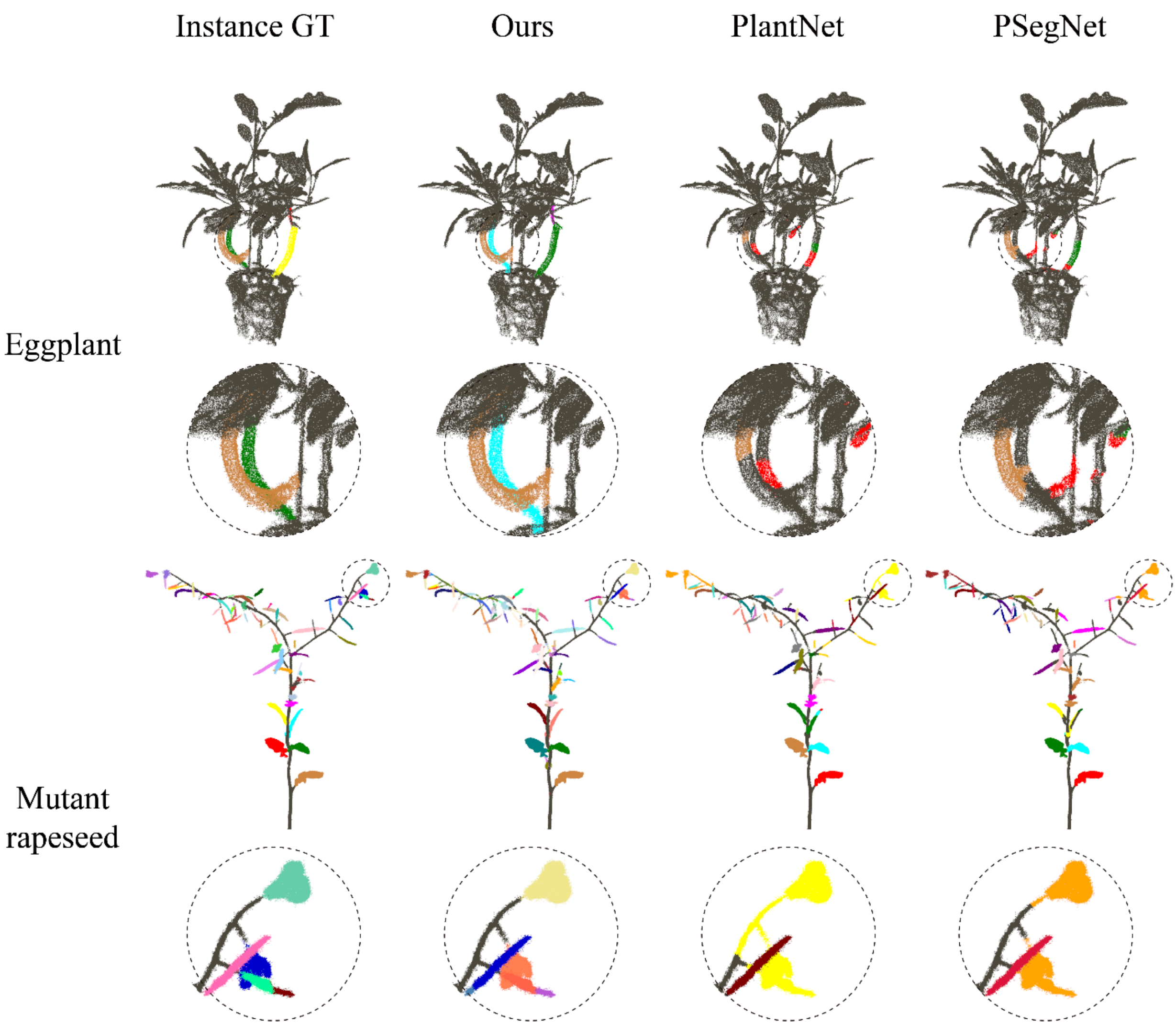

**Fig. 8.** Visualization of the instance segmentation results for six plant species datasets using the proposed PlantSegNeRF as well as PlantNet and PSegNet.

### 4.3 IM module performance

Fig. 9 demonstrates the instance matching results across six datasets, with distinct colors used to visualize different instance masks. For each representative dataset, a specific instance, labeled "Ins. A" was highlighted to show its location across different views. Immediately following YOLOv11 2D instance segmentation, the instance IDs of the same organ were inconsistent across different viewpoints. However, the instance IDs of the same organ remained consistent across the majority of multi-view images after applying the IM module, thereby demonstrating the effectiveness of the IM module. Only a few mismatches were observed, such as in the top leaves of tobacco and tomato plants.

Table 2 presents the quantitative evaluation results of instance segmentation using the IM module versus semantic clustering after plant semantic NeRF across six datasets, with the visualization example shown in Fig. 10. The method using the IM module outperformed the clustering approach in the majority of evaluation metrics. The IM module demonstrated superior performance in handling complex scenarios such as clustered top leaves and leaves with the varied leaf size, while clustering tended to incorrectly merge adjacent leaves into single instances. In yield-organ extraction and multi-object segmentation tasks, semantic clustering achieved comparable mPrec to the IM method under low organ overlap conditions but performed weaker in other metrics. The main errors of clustering were observed in incorrect merging of adjacent flowers and intertwined eggplants. In cases of sparse organ spatial distribution, the proposed semantic point cloud pipeline combined with spatial clustering achieved effective instance segmentation results. Furthermore, the IM method exhibited superior performance in handling of fine details, demonstrating its essential role in multi-scale plant instance segmentation tasks.

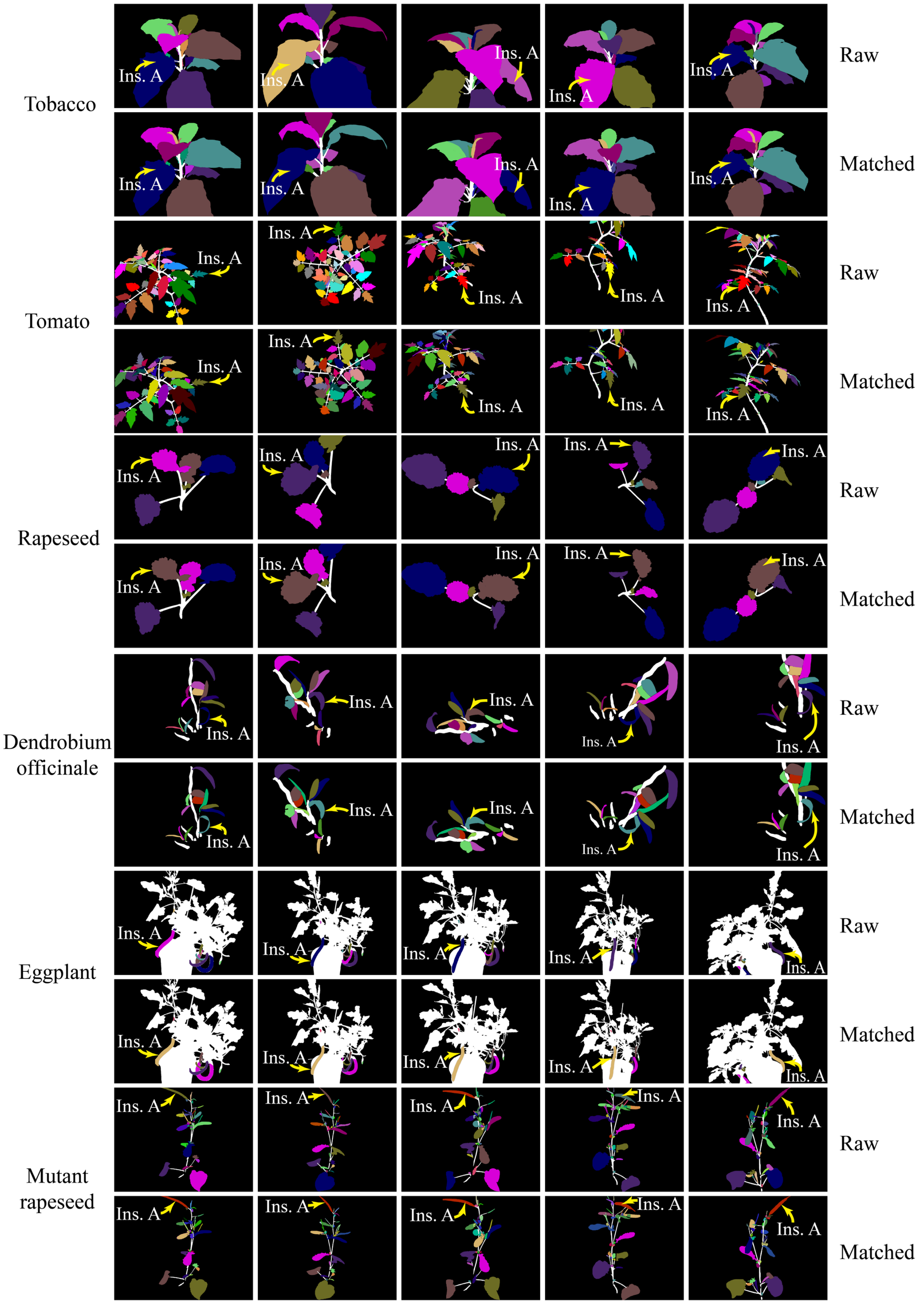

**Fig. 9.** Visualization of two-dimensional (2D) instance masks before and after instance matching (IM) across five different viewpoints, with distinct colors representing different instance identifications (IDs). One specific instance "Ins. A" is highlighted with yellow arrows to demonstrate multi-view correspondence.

**Table 2.** Comparison of instance segmentation performance with and without instance matching

(IM) module which used semantic point cloud clustering after plant semantic neural radiance fields (NeRF). The best results are in boldface.

| Plant dataset | Method | Instance segmentation metrics | | | |
|---|---|---|---|---|---|
| | | mPrec | mRec | mCov | mWCov |
| Tobacco | IM | **0.83** | **0.93** | **0.91** | **0.96** |
| | Without IM | 0.43 | 0.75 | 0.74 | 0.90 |
| Tomato | IM | **0.83** | **0.88** | **0.86** | **0.92** |
| | Without IM | 0.63 | 0.66 | 0.69 | 0.71 |
| Rapeseed | IM | **0.73** | **0.93** | **0.91** | **0.91** |
| | Without IM | 0.51 | 0.87 | 0.87 | **0.91** |
| Dendrobium officinale | IM | **0.57** | **0.91** | **0.88** | **0.89** |
| | Without IM | 0.41 | 0.50 | 0.57 | 0.61 |
| Eggplant | IM | 0.56 | **0.75** | **0.70** | **0.74** |
| | Without IM | **0.59** | 0.70 | 0.68 | 0.70 |
| Mutant rapeseed | IM | 0.74 | **0.94** | **0.88** | **0.93** |
| | Without IM | **0.78** | 0.80 | 0.81 | 0.91 |

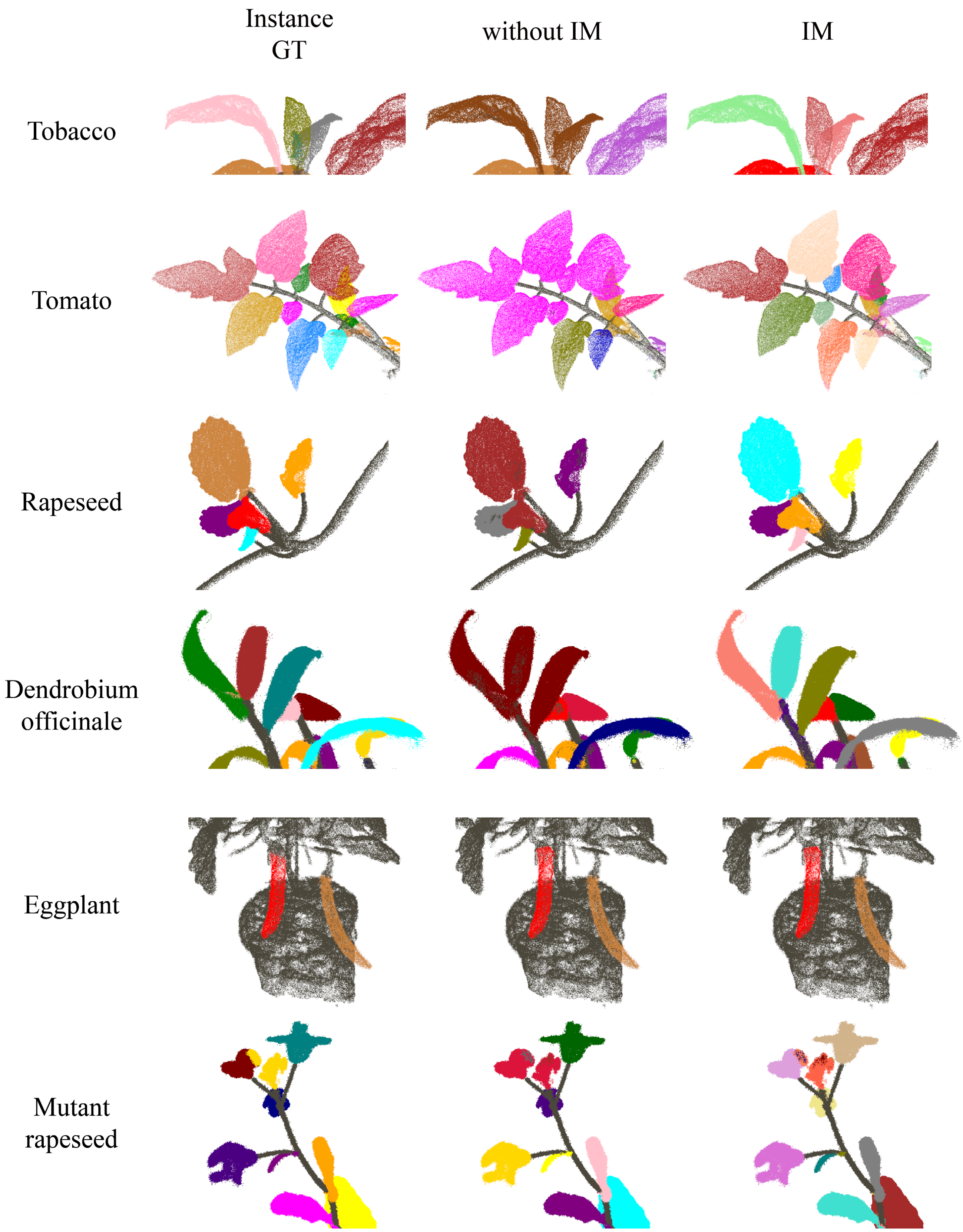

**Fig. 10.** Visualization of instance segmentation results with and without instance matching (IM) module which used semantic point cloud clustering after plant semantic neural radiance fields (NeRF).

# 5 Discussion

## 5.1 Comparisons among different methods

We proposed a novel plant segmentation approach named PlantSegNeRF that begins with 2D image instance segmentation, proceeds to match instance IDs across multiple views, and finally succeeds in the development of an instance NeRF for the reconstruction of 3D instance point clouds. PlantSegNeRF has demonstrated superior performance across all datasets, ranging from simple to complex scenarios. The cross-species superiority of PlantSegNeRF can be attributed to several key modules. It employs the YOLOv11 algorithm for the instance segmentation of 2D images. Compared to direct segmentation on 3D point clouds, which typically require downsampling to a limited number of points (e.g., 4096 points) due to computational constraints, 2D image segmentation can process high-resolution images with full detail preservation, particularly effective in capturing small organ structures. Moreover, during instance matching, PlantSegNeRF performs precise projection with occlusion relationships based on existing camera poses, which not only matches the IDs of each organ across different images but also processed images with suboptimal 2D segmentation results, including merging over-segmented instance masks and reducing the involvement of under-segmented instance masks in subsequent operations. Additionally, the proposed instance NeRF aligns instance encoding with the majority of images through ray sampling and volume rendering. This effectively filters out noise from views with poor segmentation or incorrect instance matching, providing robust support for the cross-species adaptability of our method. To facilitate the application of PlantSegNeRF on diverse plant datasets, the performance of YOLOv11 models trained for each dataset along with their corresponding point cloud instance and semantic segmentation accuracies are provided in Table S3 as reference benchmarks. Users working with different plant morphologies can refer to the image segmentation accuracies of similar datasets, noting that suboptimal 2D segmentation results will impact the final performance. Notably, PlantSegNeRF shows no significant advantages over PlantNet and PSegNet on tobacco samples in semantic segmentation. This is likely because tobacco plants have simple, upright stems situated at the central position of the plant with distinct structural boundaries, so a simple bounding box can readily outline the stem region. Nevertheless, in instance segmentation, our method presents a clear advantage as shown in Fig. 7 and Fig. 8. It is because that the young leaves at the tobacco apex are densely clustered with significant feature variations, presenting considerable segmentation challenges that other methods struggle to handle effectively, while our method can still accurately segment individual young leaves at the 2D level.

In this study, among the methods that first reconstruct the 3D point cloud and then perform segmentation, PlantNet and PSegNet consistently outperformed other methods, including PointNet++, DGCNN, and PAConv on all datasets except for tomato. Their superior performance can be attributed to their innovative architectures. PlantNet benefits from its dynamic graph convolution design, while PSegNet leverages feature fusion with an attention mechanism. However, significant performance degradation was observed across existing methods when applied to challenging datasets such as dendrobium officinale, eggplant and mutant rapeseed at podding-stage. These limitations primarily resulted from three aspects: the loss of fine-grained features due to point cloud downsampling, the challenge of organ overlap in densely growing plants, and the imbalanced morphological variations between different organs. In contrast, our method PlantSegNeRF addressed these challenges through 2D image segmentation, cross-view instance matching, and instance NeRF, achieving strong segmentation performance even for complex plant structures.

### 5.2 Impact of multi-view image number on PlantSegNeRF performance

Although multi-view imaging provides a low-cost and efficient solution for 3D reconstruction, it is known that the quality of images would affect the performance of PlantSegNeRF. Fig. 11 illustrates the relationship between PlantSegNeRF performance and the number of multi-view images, evaluated in terms of completeness, segmentation accuracy, and computational efficiency across six plant species datasets. When the number of images was insufficient, both segmentation accuracy and point cloud integrity deteriorated, or in some cases, even failed to generate the point cloud. This adverse effect can be attributed to two main factors: reduced viewpoints participating in voting increased the difficulty of instance matching, while limited viewpoints led to poor network convergence in NeRF training, resulting in errors of volume density, color, as well as semantic and instance encoding. As the number of images increases, IoU, mWCov, and completeness rose rapidly beyond a minimum camera threshold, and then remained stable with only minor fluctuations. In terms of processing speed, the computational time increased approximately linearly with the number of images. Plants with relatively simple morphology, such as tobacco, tomato and rapeseed, required less processing time than those with more complex structures. Mutant rapeseed at the podding stage and eggplant at the fruit stage required 70 and 90 images, respectively, to achieve stable performance, while the relatively simple datasets reached stability with as few as 50 images. The low image requirement of the PlantSegNeRF method demonstrates that RGB cameras can capture sufficient data for 3D reconstruction and organ segmentation in a short period. For six-camera systems like the one used in this study, data acquisition can be completed within seconds, providing a solid foundation for large-scale, high-throughput plant phenotyping.

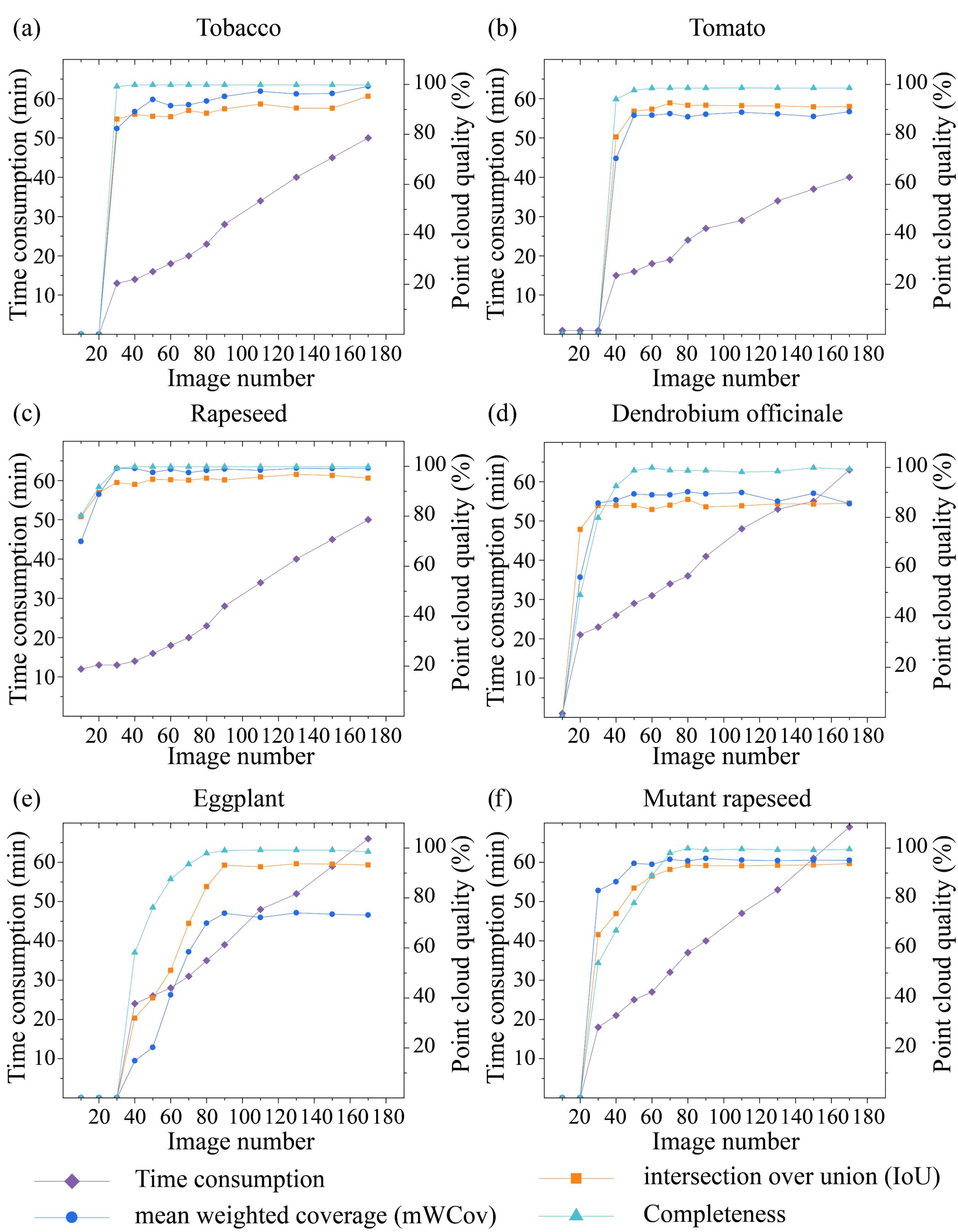

**Fig. 11.** The impact of multi-view image number on PlantSegNeRF performance in terms of time consumption, intersection over union (IoU), mean weighted coverage (mWCov) and completeness across six plant species datasets: (a) tobacco, (b) tomato, (c) rapeseed, (d) dendrobium officinale, (e) eggplant, and (f) mutant rapeseed.

### 5.3 Dependency of plant training dataset

Fig. 12 and Fig. 13 present the instance and semantic segmentation performance of different methods under limited plant specimen training conditions, where the sample number refers to the actual number of plant individuals. It was found that while the segmentation performance of other methods improved with an increasing size of the training plant dataset, the rate of improvement gradually levels off. Except for the semantic segmentation task on the structurally simple tobacco dataset, all other datasets exhibited significantly lower segmentation performance than PlantSegNeRF. Even with 40 training samples, methods that perform segmentation on point cloud struggled to achieve the effectiveness that our method accomplished with only two plant samples. This performance gap may be attributed to two main factors. First, the use of 2D multi-view images for annotation and training allows a large number of diverse and representative views to be obtained from just one or two plant samples. Second, the YOLO 2D segmentation model, pre-trained on extensive image datasets, possesses strong feature extraction capabilities. This enables the model to effectively learn plant characteristics during fine-tuning with limited local samples and to generalize well across similar plants. In addition, our proposed IM module can be integrated to further correct suboptimal 2D segmentation results, thereby improving overall segmentation accuracy. The minimal dependence on extensive plant sample training data opens up a new opportunity in plant phenotyping. Researchers only need to collect multi-view images of a few individual plants to train a segmentation model applicable to specific plant variety, which significantly reduces the initial data collection workload, making high-throughput phenotyping more accessible and feasible.

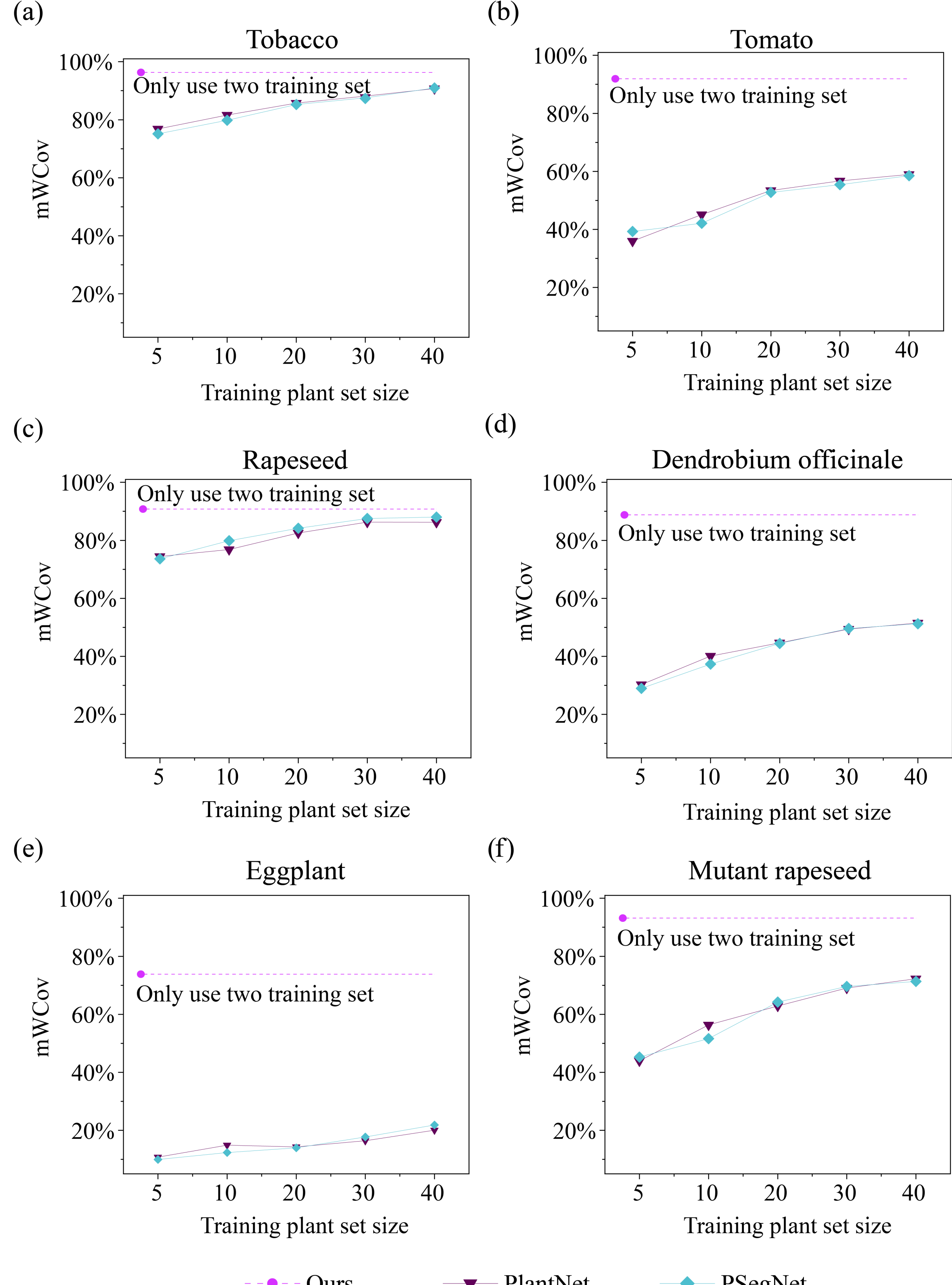

**Fig. 12.** Point cloud instance segmentation performance among the proposed PlantSegNeRF, PlantNet, and PSegNet under different numbers of training plant samples across six plant species datasets: (a) tobacco, (b) tomato, (c) rapeseed, (d) dendrobium officinale, (e) eggplant, and (f) mutant rapeseed.

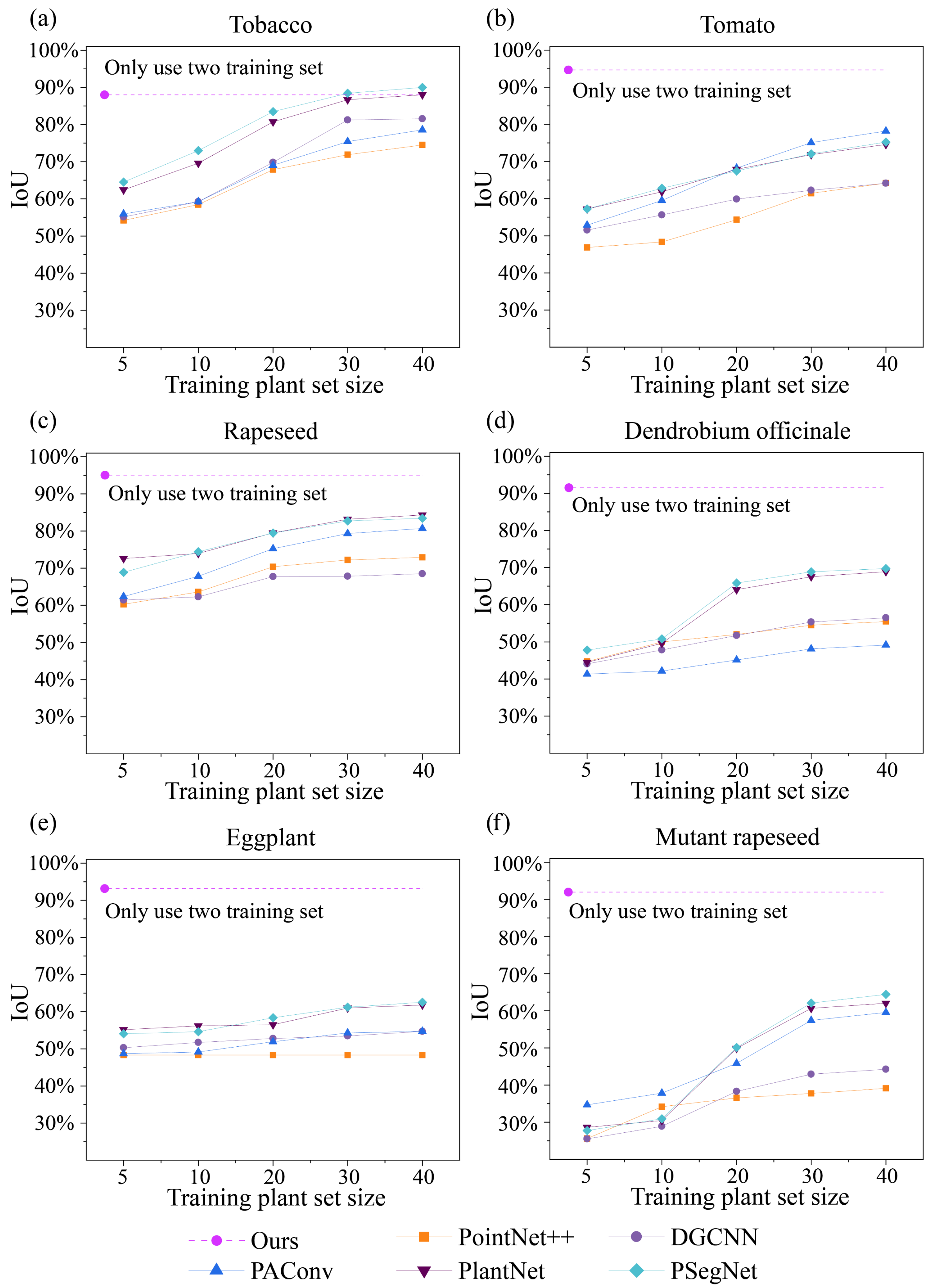

**Fig. 13.** Point cloud semantic segmentation performance between the proposed PlantSegNeRF as well as PointNet++, DGCNN, PAConv, PlantNet, and PSegNet under different numbers of training plant samples across six plant species datasets: (a) tobacco, (b) tomato, (c) rapeseed, (d) dendrobium officinale, (e) eggplant, and (f) mutant rapeseed.

### 5.4 Limitations and Prospects

This study presents a novel framework for generating high-precision plant instance point clouds from multi-view images using IM module and instance NeRF. PlantSegNeRF itself can generate large-scale, instance- and semantically-annotated 3D point clouds, providing the research community with a vital resource for training next-generation 3D segmentation networks. Nevertheless, certain aspects require further refinement to achieve broader applicability in plant phenotyping. The main objective of our work is to provide assistance for downstream phenotyping tasks in greenhouse and indoor scenarios. However, wind-induced plant movement in open-field conditions would directly and adversely affect the NeRF model training. We believe that there are two main potential approaches to address wind-related challenges in future work, including increasingly advanced dynamic NeRF and 3DGS technologies, as well as multi-camera phenotyping technologies such as synchronized capture systems with 36 cameras. Future work on development of 2D segmentation foundation models, coupled with expanding high-quality plant segmentation datasets, will enable zero-shot annotation capabilities and make the proposed method widely accessible across various communities from computer science to plant studies.

# 6 Conclusions

In this study, a novel few-shot and cross-species method, called PlantSegNeRF, was proposed to generate high-precision plant instance point clouds directly from multi-view RGB image sequences for a wide range of plant species. It was found that the IM module effectively unified instance IDs across different viewpoints and significantly improved the accuracy of point cloud instance segmentation compared to the semantic clustering approach, especially in complex scenarios with clustered or overlapping organs. For semantic point cloud segmentation, PlantSegNeRF outperformed the commonly used methods, demonstrating an average improvement of 16.1%, 18.3%, 17.8%, and 24.2% in precision, recall, F1-score, and IoU compared to the second-best results on structurally complex datasets. More importantly, PlantSegNeRF exhibited significant advantages in plant point cloud instance segmentation tasks. Across all plant datasets, PlantSegNeRF achieved average improvements of 11.7%, 38.2%, 32.2% and 25.3% in mPrec, mRec, mCov, and mWCov, respectively, compared to the second-best method, PSegNet. These substantial gains are attributed to the combination of accurate 2D instance segmentation, robust instance matching across multiple views, and effective noise filtering during NeRF training, which together enhance the performance of PlantSegNeRF across different datasets. Furthermore, PlantSegNeRF has demonstrated superior few-shot performance, as it only needs to collect multi-view images of a few individual plants to train a segmentation model applicable to a specific plant variety, which significantly reduces the initial data collection workload, making high-throughput phenotyping more accessible and feasible.

Further research is needed to leverage more advanced large models to replace the YOLOv11 module for 2D image segmentation, which would make the approach more user-friendly and accessible. This aligns with the growing interest in zero-shot annotation, which is highly beneficial for phenotyping applications. From another perspective, the PlantSegNeRF method can generate a large number of plant 3D point clouds with high-precision instance and semantic information. This is crucial for developing large-scale, high-accuracy segmentation of 3D point cloud modules, thus serving as a foundational data resource.

# Author contributions

X.Y., R.D., H.H., J.X., P.X., L.F. and Z.G. designed the study, performed the experiment and wrote the manuscript. X.Y. developed the algorithm and analyzed the data. N.J. guided the dataset design and provided plant samples. Y.J. reviewed and revised the manuscript. H.C. supervised experiments at all stages and carried out revisions of the manuscript.

# Funding

This work was supported by the International S&T Cooperation Program of China (2024YFE0115000), National Natural Science Foundation of China (32371985), Amway (China) Enterprise Project (Am20210428RD), and Joint Laboratory of Zhejiang University and China Tobacco Zhejiang Industrial Co., LTD (2023-KYY-513108-0011).

# Declaration of Competing Interest

The authors declare that there is no conflict of interest regarding the publication of this article.

# Acknowledgements

This work was supported by the Earth System Big Data Platform of the School of Earth Sciences, Zhejiang University.

# Supplementary Materials

**Table S1.** Detailed parameter configurations of models utilized in this study

| Model | Parameter |
|---|---|
| You Only Look Once version 11 (YOLOv11) | Image size: 2160 |
| | Training set to testing set ratio: 8:2 |
| | Initial learning rate: 0.01 |
| | Final learning rate: 0.0001 |
| | Optimizer: SGD |
| | Batch size: 16 |
| | Epoch number: 100 |
| | Parameters size: 3.4 M |
| | Model file weight: 6.45 MB |
| | Inference time: 11.3 ms |
| Instance Neural Radiance Fields (NeRF) | Image resolution: original |
| | Training set to testing set ratio: 9:1 |
| | Optimizer: Adam |
| | Rays per iteration: 2048 |
| | Number of iterations: 10,000 |
| | Near plane sampling bound: 0.05 |
| | Far plane sampling bound: 1000 |
| | Coarse sampling points per ray: 64 |
| | Fine sampling points per ray: 128 |
| | Occupancy grid resolution (ref. Instant-NGP): $128^3$ |
| | Early ray termination threshold (ref. Instant-NGP): 0.99 |
| | Point cloud extraction: 1,000,000 |

**Table S2.** Specifications of key computational hardware components used in this study

| Hardware component | Model |
|---|---|
| Server architecture | MSI X570-A PRO |
| GPU | NVIDIA GeForce RTX 4090 (24 GB) |
| CPU | AMD 5900 |
| RAM | 128 GB (3600 MT/s) |

**Table S3.** Performance benchmarks of YOLOv11 models and corresponding point cloud segmentation accuracies across different plant datasets, with mean Average Precision at IoU threshold 0.5 for Main images (mAP50(M)) for two-dimensional (2D) image segmentation, Intersection over Union (IoU) and mean Weighted Coverage (mWCov) representing semantic and instance segmentation respectively.

| **Plant dataset** | **2D image segmentation** | **Point cloud semantic segmentation** | **Point cloud instance segmentation** |
|---|---|---|---|
| | mAP50(M) | IoU | mWCov |
| Tobacco | 87.4 | 88.0 | 96.3 |
| Tomato | 85.8 | 94.7 | 91.9 |
| Rapeseed | 97.3 | 95.0 | 90.8 |
| Dendrobium officinale | 82.6 | 91.5 | 88.8 |
| Eggplant | 80.9 | 93.1 | 73.9 |
| Mutant rapeseed | 78.7 | 92.0 | 93.2 |

**Equation S1**

The adaptive resolution minimum rectangle sampling strategy aims to achieve more balanced sampling of projection points (Fig. 2b), for example, sampling more points per unit area for small leaves and more points along the short-edge direction for elongated leaves. Specifically, when performing point sampling on a 2D image, the minimum bounding rectangle is first computed for each instance (e.g., each leaf), then the minimum bounding rectangle is used as the coordinate system and the sampling density along the long-edge and short-edge directions is adaptively set based on the length and width of the bounding rectangle. The sampling step size is determined based on the edge length of the minimum bounding rectangle, as described in Equation (S1):

$$s = \begin{cases} 4 & (d \leq 40) \\ \left\lfloor \frac{d}{10} \right\rfloor & (40 < d \leq 400) \\ 40 & (d > 400) \end{cases} \tag{S1}$$

where $s$ represents the sampling step size in pixels, $d$ represents the edge length of the rectangle in pixels. The sampling step sizes for the long edge and short edge are calculated independently.

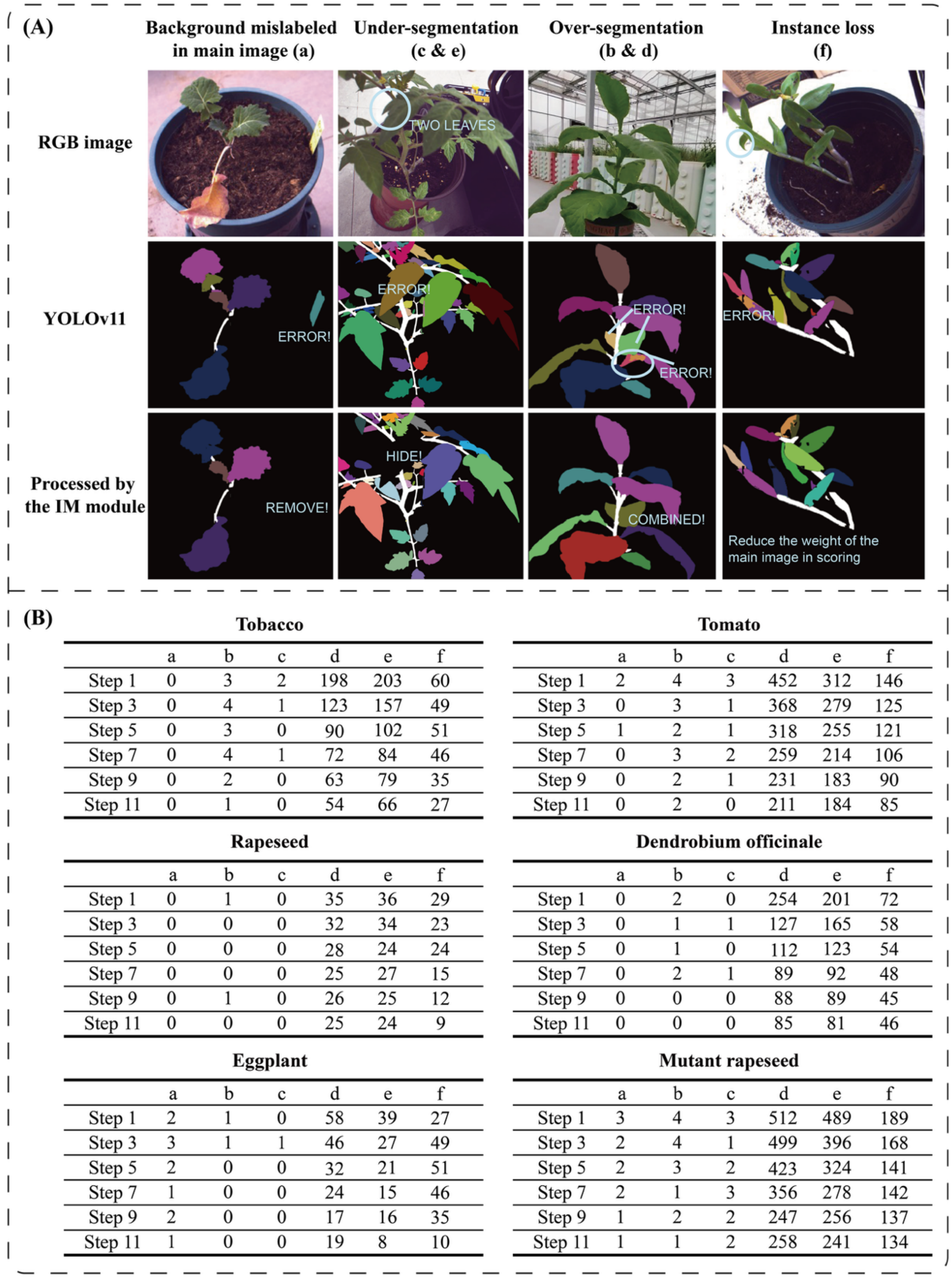

**Tobacco**

| | a | b | c | d | e | f |
|---|---|---|---|---|---|---|
| Step 1 | 0 | 3 | 2 | 198 | 203 | 60 |
| Step 3 | 0 | 4 | 1 | 123 | 157 | 49 |
| Step 5 | 0 | 3 | 0 | 90 | 102 | 51 |
| Step 7 | 0 | 4 | 1 | 72 | 84 | 46 |
| Step 9 | 0 | 2 | 0 | 63 | 79 | 35 |
| Step 11 | 0 | 1 | 0 | 54 | 66 | 27 |

**Tomato**

| | a | b | c | d | e | f |
|---|---|---|---|---|---|---|
| Step 1 | 2 | 4 | 3 | 452 | 312 | 146 |
| Step 3 | 0 | 3 | 1 | 368 | 279 | 125 |
| Step 5 | 1 | 2 | 1 | 318 | 255 | 121 |
| Step 7 | 0 | 3 | 2 | 259 | 214 | 106 |
| Step 9 | 0 | 2 | 1 | 231 | 183 | 90 |
| Step 11 | 0 | 2 | 0 | 211 | 184 | 85 |

**Rapeseed**

| | a | b | c | d | e | f |
|---|---|---|---|---|---|---|
| Step 1 | 0 | 1 | 0 | 35 | 36 | 29 |
| Step 3 | 0 | 0 | 0 | 32 | 34 | 23 |
| Step 5 | 0 | 0 | 0 | 28 | 24 | 24 |
| Step 7 | 0 | 0 | 0 | 25 | 27 | 15 |
| Step 9 | 0 | 1 | 0 | 26 | 25 | 12 |
| Step 11 | 0 | 0 | 0 | 25 | 24 | 9 |

**Dendrobium officinale**

| | a | b | c | d | e | f |
|---|---|---|---|---|---|---|
| Step 1 | 0 | 2 | 0 | 254 | 201 | 72 |
| Step 3 | 0 | 1 | 1 | 127 | 165 | 58 |
| Step 5 | 0 | 1 | 0 | 112 | 123 | 54 |
| Step 7 | 0 | 2 | 1 | 89 | 92 | 48 |
| Step 9 | 0 | 0 | 0 | 88 | 89 | 45 |
| Step 11 | 0 | 0 | 0 | 85 | 81 | 46 |

**Eggplant**

| | a | b | c | d | e | f |
|---|---|---|---|---|---|---|
| Step 1 | 2 | 1 | 0 | 58 | 39 | 27 |
| Step 3 | 3 | 1 | 1 | 46 | 27 | 49 |
| Step 5 | 2 | 0 | 0 | 32 | 21 | 51 |
| Step 7 | 1 | 0 | 0 | 24 | 15 | 46 |
| Step 9 | 2 | 0 | 0 | 17 | 16 | 35 |
| Step 11 | 1 | 0 | 0 | 19 | 8 | 10 |

**Mutant rapeseed**

| | a | b | c | d | e | f |
|---|---|---|---|---|---|---|
| Step 1 | 3 | 4 | 3 | 512 | 489 | 189 |
| Step 3 | 2 | 4 | 1 | 499 | 396 | 168 |
| Step 5 | 2 | 3 | 2 | 423 | 324 | 141 |
| Step 7 | 2 | 1 | 3 | 356 | 278 | 142 |
| Step 9 | 1 | 2 | 2 | 247 | 256 | 137 |
| Step 11 | 1 | 1 | 2 | 258 | 241 | 134 |

**Fig. S1.** IM module processing details. (A) Visualization of error examples in two-dimensional (2D) segmentation and corresponding solutions using instance matching (IM) module. (B) Number of different error types identified by the algorithm across six species datasets over iterations. a, b, c, d, e, f represent background mislabeled in main image, over-segmentation in main image, under-segmentation in main image, over-segmentation in auxiliary images, under-segmentation in auxiliary images, and instance loss, respectively.

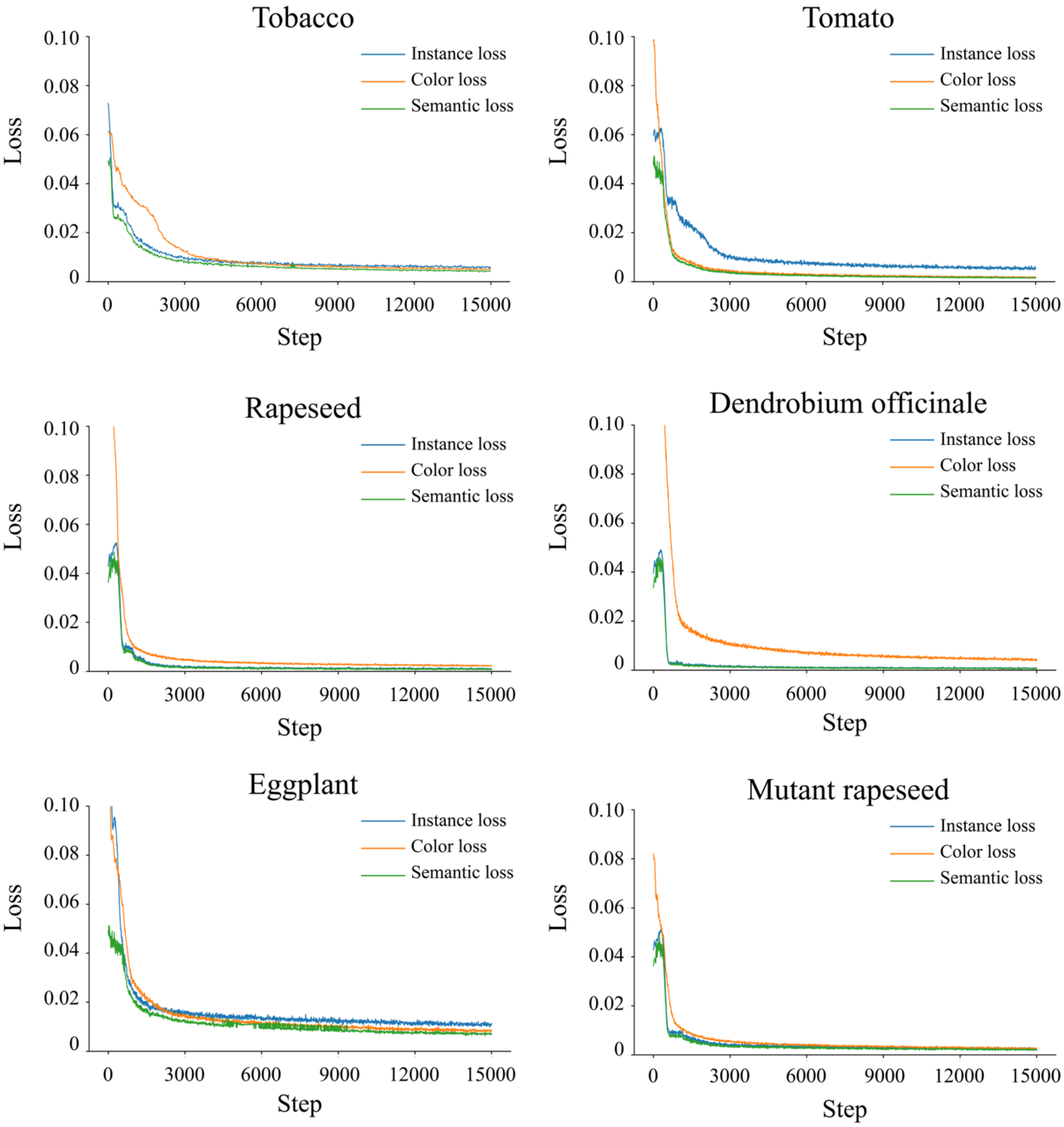

**Fig. S2.** Visualization of representative loss function components during instance neural radiance fields (NeRF) training across six plant species datasets, with loss values normalized by dividing by 255.